%% file: main.tex
\documentclass[10pt]{article} 
\usepackage[preprint]{tmlr}

\input{math_commands.tex}

\usepackage{hyperref}
\usepackage{url}

\usepackage{amsmath,amsthm,amssymb}

\usepackage{mathtools}
\usepackage{subcaption}
\usepackage{enumitem}
\usepackage{algorithm}
\usepackage{algorithmic}
\usepackage{booktabs}
\usepackage{multirow}
\usepackage{bbm}
\usepackage{adjustbox}

\numberwithin{equation}{section}
\newtheorem{theorem}{Theorem}[section]
\newtheorem{proposition}[theorem]{Proposition}

\theoremstyle{assumption}

\theoremstyle{definition}

\theoremstyle{remark}

\title{\texorpdfstring{$\phi$}{phi}-Table: A Statistical Explanation for Global SHAP}

\author{\name Dongseok Kim\thanks{These authors contributed equally.} \email jkds5920@gachon.ac.kr \\
      \addr Department of Computer Engineering\\
      Gachon University
      \AND
      \name Hyoungsun Choi\footnotemark[1] \email hschoi@gachon.ac.kr \\
      \addr Department of Computer Engineering\\
      Gachon University
      \AND
      \name Mohamed Jismy Aashik Rasool\footnotemark[1] \email aashikrasool@gachon.ac.kr \\
      \addr Department of Computer Engineering\\
      Gachon University
      \AND
      \name Gisung Oh\thanks{Corresponding author.} \email eustia@gachon.ac.kr \\
      \addr Department of Computer Engineering\\
      Gachon University}

\def\month{MM}  
\def\year{YYYY} 
\def\openreview{\url{https://openreview.net/forum?id=XXXX}} 

\begin{document}

\maketitle

\begin{abstract}
Global SHAP explanations are typically presented as feature-importance rankings, which identify variables that matter to a black-box model but do not indicate whether their effects admit clear directional summaries, how uncertain those summaries are, or how faithfully they represent the fitted response. This paper proposes the $\phi$-table, a SHAP-based statistical explanation table for tabular black-box regression models. The procedure selects features by SHAP importance and fits a standardized linear surrogate to the fitted model response $f(X)$, reporting SHAP importance together with model-response coefficients, uncertainty summaries, surrogate fidelity, and bootstrap coefficient stability. The resulting coefficients are interpreted as projections of the fitted model response onto the SHAP-selected feature set. Across synthetic, semi-synthetic, and real-data experiments, the $\phi$-table extends ranking-only SHAP into a statistical global explanation by exposing direction, uncertainty, fidelity, and stability as distinct components of fitted model behavior.
\end{abstract}

\section{Introduction}
\label{sec:introduction}

Global feature attribution methods are widely used to summarize the behavior of black-box predictive models. Among them, SHAP has become a standard tool for tabular model explanation because local attributions can be aggregated into a global feature-importance ranking. Such a ranking is useful: it tells the analyst which variables contribute most strongly to the fitted model response in attribution magnitude. Yet a ranking-only explanation leaves several statistical questions unanswered. It does not indicate whether an important feature admits a clear positive or negative directional summary, how uncertain that direction is, whether the directional summary is stable under resampling, or how faithfully the selected features approximate the black-box response.

These questions matter because importance and direction are distinct objects. A feature can be important to a black-box model through nonlinear effects, symmetric responses, interactions, thresholds, or region-specific behavior, while having no stable global direction in a low-dimensional summary. Conversely, a feature with a clear directional coefficient need not be the most important feature by attribution magnitude. Treating global importance as if it were already a directional explanation can therefore lead to overinterpretation. A more informative global explanation should preserve the model-aware screening role of SHAP while making direction, uncertainty, approximation quality, and stability explicit.

This work proposes the $\phi$-table, a SHAP-based statistical explanation table for tabular black-box regression models. The procedure first uses mean absolute SHAP values to select the top-ranked features. Conditional on this selected feature set, it then fits a standardized linear surrogate to the fitted black-box response $f(X)$, rather than to the original outcome $Y$. The resulting table reports SHAP importance together with surrogate coefficients, robust standard errors, confidence intervals, $p$-values, surrogate fidelity, and bootstrap coefficient stability. Its coefficients are therefore summaries of trained model behavior over the explanation sample: they describe model-response projections, not causal effects or structural parameters of the data-generating process.

The main contributions are as follows:
\begin{itemize}
    \item \textbf{A statistical reporting layer for global SHAP.}
    The $\phi$-table preserves SHAP as the feature-selection and importance mechanism, but changes the reporting object from a ranking alone to a table that jointly presents attribution magnitude, model-response direction, uncertainty, fidelity, and coefficient stability.

    \item \textbf{Separation of importance and direction.}
    The analysis formalizes why high attribution magnitude need not imply a stable positive or negative model-response coefficient.

    \item \textbf{Model-response projection target.}
    Surrogate coefficients are defined as linear projections of the fitted black-box response $f(X)$ onto the SHAP-selected covariates.

    \item \textbf{Conditional uncertainty summaries.}
    Robust standard errors, confidence intervals, and p-values are reported as nominal summaries of coefficient-level uncertainty within the SHAP-selected model-response surrogate.

    \item \textbf{Fidelity- and stability-conditioned interpretation.}
    The table is interpreted through both surrogate fidelity and bootstrap coefficient stability, rather than through coefficient signs alone.
\end{itemize}

\section{Related Work}
\label{sec:related_work}

\subsection{Feature Attribution and Global SHAP Explanations}
\label{subsec:feature_attribution_global_shap}

Feature attribution explains black-box predictions by assigning contribution scores to input features. Model-agnostic methods define these scores through input influence, perturbation behavior, local approximation, or cooperative-game-theoretic allocation~\citep{strumbelj2010efficient, datta2016algorithmic, ribeiro2016should, lundberg2017unified, ribeiro2018anchors}. Neural-network attribution methods use gradients, activation differences, perturbation masks, and localization maps to produce the same feature-contribution format~\citep{sundararajan2017axiomatic, shrikumar2017learning, ancona2017towards, fong2017interpretable, selvaraju2017grad}. Evaluation work further shows that attribution scores should be assessed through sensitivity, faithfulness, and robustness rather than visual plausibility alone~\citep{adebayo2018sanity, yeh2019fidelity, covert2021explaining}.

SHAP placed feature attribution in a unified additive framework based on Shapley values, and efficient algorithms for tree-based models made it especially influential for tabular data analysis~\citep{lundberg2020local}. In global SHAP explanations, local attributions are commonly aggregated by averaging absolute values over an explanation sample, producing a feature-importance ranking. Related Shapley-based work studies global importance through predictive power, global sensitivity analysis, and structured dependency representations~\citep{covert2020understanding, song2016shapley, wang2021shapley}, while later extensions address interaction attribution and faster estimation~\citep{sundararajan2020shapley, jethani2021fastshap, tsai2023faith}. These developments enrich attribution methods, but the common global reporting format remains a ranking by attribution magnitude rather than a statistical summary of direction, uncertainty, fidelity, and stability.

\subsection{Surrogate Models for Interpretable Model Summaries}
\label{subsec:surrogate_explanations}

Surrogate modeling explains a complex predictor by approximating it with a simpler model. Early rule- and tree-extraction methods learned comprehensible structures from trained neural networks, while model compression transferred predictive behavior from complex models into smaller interpretable predictors~\citep{craven1995extracting, bucilua2006model, tan2018learning, lakkaraju2019faithful}. In this setting, the explanatory target is the behavior of the fitted model rather than the data-generating process.

Transparent model classes provide another route to global model summaries. Generalized additive models and interaction-augmented variants retain feature-wise readability while allowing nonlinear shape functions~\citep{lou2012intelligible, lou2013accurate, caruana2015intelligible}. Sparse rule ensembles and scoring systems express predictions through short rules, integer scores, or compact risk tables~\citep{friedman2008predictive, ustun2016supersparse, ustun2019learning}. Rule lists, rule sets, Boolean rules, and optimal sparse trees further combine explicit structural simplicity with optimization-based learning procedures~\citep{letham2015interpretable, wang2015falling, yang2017scalable, angelino2018learning, wang2017bayesian, dash2018boolean, hu2019optimal}.

Recent methods also learn interpretable substitutes, companions, or competitors for regions where simpler structures approximate a black box well~\citep{wang2018multi, wang2019gaining, rafique2020transparency, lin2020generalized}. The $\phi$-table follows this surrogate tradition by treating the fitted response as the explanation target, but uses SHAP to select the summarized variables before fitting a standardized linear projection. The resulting table keeps feature importance and surrogate coefficient direction as separate quantities.

\subsection{Statistical Uncertainty in Model Explanations}
\label{subsec:uncertainty_explanations}

Feature attributions and surrogate summaries are estimates whose values can vary with perturbation draws, background samples, auxiliary models, fitted predictors, or explanation samples. Uncertainty-aware explanation methods make this variability explicit through Bayesian uncertainty for local explanations, bootstrap uncertainty for feature-importance models, or uncertainty-aware Shapley-value estimation~\citep{slack2021reliable, schwab2019cxplain, covert2021improving, watson2023explaining}. Statistical inference for feature importance similarly defines population-level importance targets and develops confidence intervals or hypothesis tests for them~\citep{williamson2020efficient, williamson2021nonparametric, williamson2023general, van2006statistical}. Related Rashomon-set work shows that equally accurate models may support different explanatory patterns~\citep{fisher2019all, laberge2023partial}.

The coefficient columns of the $\phi$-table connect explanation uncertainty to regression-style inference. Conditional on the SHAP-selected feature set, the table fits a standardized surrogate regression to the fitted response and reports standard errors, confidence intervals, and $p$-values for model-response coefficients. Robust covariance estimators and resampling methods provide tools for quantifying finite-sample variability under heteroskedasticity or limited analytic justification~\citep{white1980heteroskedasticity, efron1992bootstrap}. Because the feature set is selected before the coefficient table is fit, the procedure also relates to post-selection and selective inference~\citep{berk2013valid, lee2016exact, lockhart2014significance, yang2016selective, suzumura2017selective}. Predictive uncertainty provides a broader connection, since fitted model responses may vary because of epistemic uncertainty, aleatoric uncertainty, or model instability~\citep{gal2016dropout, kendall2017uncertainties, lakshminarayanan2017simple}.

\subsection{Reliability, Fidelity, and Stability of Explanations}
\label{subsec:reliability_explanations}

Reliability-oriented work evaluates whether explanations faithfully and reproducibly reflect the model being explained. Saliency maps and feature attributions can change under small perturbations, input transformations, hyperparameter choices, or model differences~\citep{alvarez2018towards, ghorbani2019interpretation, bansal2020sam, shah2021input}. Benchmark studies translate these concerns into quantitative evaluation protocols, including remove-and-retrain tests, controlled feature-importance benchmarks, time-series interpretability benchmarks, and human-centered evaluations~\citep{hooker2019benchmark, ismail2020benchmarking, jeyakumar2020can, nguyen2021effectiveness}. Evaluation toolkits and surveys organize these criteria around faithfulness, robustness, localization, complexity, stability, and user-grounded effectiveness~\citep{agarwal2022openxai, hedstrom2023quantus, nauta2023anecdotal}.

Faithfulness and fidelity are especially important for surrogate explanations because a surrogate is useful only insofar as it captures the relevant behavior of the original model. Existing work studies whether local explanations match the predictor, whether importance scores identify features that truly affect predictions, and whether faithfulness metrics are themselves reliable~\citep{dasgupta2022framework, madsen2022evaluating, madsen2023faithfulness}. Robustness adds another dimension: smoothing, invariance constraints, or robust training can reduce sensitivity without necessarily preserving faithfulness~\citep{crabbe2023evaluating, tan2023robust}. Since explanations can also be manipulated or stabilized without changing predictions, recent work studies robustness constraints, stable-explanation training, and meta-evaluation procedures~\citep{dombrowski2019explanations, wicker2022robust, chen2024training, hedstrom2023meta}. The $\phi$-table incorporates this perspective through surrogate fidelity for approximation quality and bootstrap coefficient stability for resampling reproducibility.

\section{Method}
\label{sec:method}

\subsection{Model-Response Explanation}
\label{subsec:model_response_explanation}

Let $\mathcal{D}=\{(x_i,y_i)\}_{i=1}^n$ denote a tabular regression dataset, where $x_i\in\mathbb{R}^d$ is a vector of covariates and $y_i$ is the observed response. A black-box model is trained on this data to produce a fitted prediction function $f:\mathbb{R}^d\rightarrow\mathbb{R}$. Our goal is to explain the fitted model response over an empirical covariate distribution, rather than to estimate the data-generating relationship between $X$ and $Y$.

For an explanation sample $\mathcal{E}=\{x_i\}_{i=1}^{m}$, we write
\[
    z_i = f(x_i),
    \qquad i=1,\ldots,m .
\]
The statistical object summarized by the proposed table is therefore $z_i$, not the observed response $y_i$. This model-response perspective separates the outcome used to train the black-box model from the fitted response subsequently explained by the table. The reported coefficients, intervals, and diagnostic quantities should consequently be read as summaries of trained model behavior over the explanation sample.

A useful global explanation should not stop at identifying influential variables. It should also indicate whether those variables admit a directional summary, how uncertain that summary is, and how well the resulting low-dimensional table approximates the black-box response. The following subsections describe how SHAP values select the variables to be summarized and how the selected variables are converted into the proposed $\phi$-table.

\subsection{SHAP-Based Feature Selection}
\label{subsec:shap_feature_selection}

Given the fitted model $f$, we compute SHAP values on the explanation sample $\mathcal{E}$. Let $\varphi_{ij}$ denote the SHAP value of feature $j$ for observation $x_i$. We aggregate these local attributions into a global importance score by taking their mean absolute magnitude:
\[
    I_j
    =
    \frac{1}{m}\sum_{i=1}^{m} |\varphi_{ij}|,
    \qquad j=1,\ldots,d .
\]
Features are ranked in decreasing order of $I_j$. For a chosen table size $k$, let
\[
    S_k
    =
    \{j_1,\ldots,j_k\}
\]
denote the indices of the top-$k$ features under this ranking.

In our procedure, SHAP serves as a model-aware screening mechanism. It determines which variables enter the explanation table by identifying features with large attribution magnitude for the fitted response $f(X)$. It does not, by itself, assign a single global direction or quantify coefficient-level uncertainty. This distinction matters because a feature may be important for prediction through nonlinear effects, interactions, or region-specific behavior, even when it has no stable positive or negative linear summary over the explanation sample.

The selected set $S_k$ is therefore treated as the input to a second-stage statistical summary. Conditional on this set, we fit a low-dimensional surrogate model to the black-box response and report the resulting coefficient-level quantities in the $\phi$-table. This two-stage construction is deliberate: SHAP determines which variables enter the global explanation, while the surrogate table determines what statistical claims can be made about those variables as summaries of $f(X)$.

\subsection{Constructing the \texorpdfstring{$\phi$}{phi}-Table}
\label{subsec:constructing_phi_table}

Given the SHAP-selected feature set $S_k$, we construct the table by projecting the fitted model response onto the selected covariates. Let $\tilde z_i$ denote the standardized version of $z_i=f(x_i)$, and let $\tilde x_{ij}$ denote the standardized value of selected feature $j\in S_k$. The surrogate model is
\[
    \tilde z_i
    =
    \alpha
    +
    \sum_{j\in S_k} \beta_j \tilde x_{ij}
    +
    \varepsilon_i,
    \qquad i=1,\ldots,m .
\]
This regression provides a linear summary of the trained model response using only the SHAP-selected variables.

The coefficient $\hat\beta_j$ summarizes the direction of feature $j$ within this selected-feature surrogate. A positive coefficient means that larger values of feature $j$ are associated with larger fitted model responses, conditional on the other selected covariates in the surrogate. A negative coefficient indicates the opposite direction. Since both the surrogate response and covariates are standardized, coefficient magnitudes are comparable across selected features within the same table.

For each selected feature, the \(\phi\)-table reports the feature name, SHAP importance \(I_j\), SHAP rank, surrogate coefficient \(\hat\beta_j\), standard error, confidence interval, and two-sided \(p\)-value. These uncertainty columns are interpreted conditionally on the SHAP-selected feature set and on the selected-feature surrogate target. In the experiments, coefficient uncertainty is computed using HC3 heteroskedasticity-robust standard errors. If \(\widehat{\mathrm{se}}_{\mathrm{HC3}}(\hat\beta_j)\) denotes the HC3 standard error, the nominal confidence interval is
\[
\hat\beta_j \pm t_{m-k-1, 1-\gamma/2}\widehat{\mathrm{se}}_{\mathrm{HC3}}(\hat\beta_j),
\]
where \(1-\gamma\) is the nominal confidence level.

The resulting table combines two complementary summaries. SHAP importance records how strongly a feature contributes to the black-box prediction in attribution magnitude, while the surrogate coefficient gives a directional model-response summary for the selected feature. The confidence interval and \(p\)-value summarize nominal uncertainty associated with this coefficient within the selected-feature surrogate. They are not selection-adjusted guarantees for the full SHAP ranking-and-selection procedure. In this way, the \(\phi\)-table extends a ranking-only global explanation into a statistical explanation table.

\subsection{Diagnostics and Interpretive Scope}
\label{subsec:diagnostics_and_scope}

The coefficient columns are reported together with table-level diagnostics. The first diagnostic is surrogate fidelity, which measures how well the selected-feature surrogate reproduces the fitted black-box response on the explanation sample. Let \(\hat z_i^{\,\mathrm{sur}}\) denote the fitted value from the surrogate. We define
\[
    \mathrm{Fid}
    =
    1
    -
    \frac{
        \sum_{i=1}^{m}
        (\tilde z_i-\hat z_i^{\,\mathrm{sur}})^2
    }{
        \sum_{i=1}^{m}
        (\tilde z_i-\bar{\tilde z})^2
    } .
\]
Higher fidelity indicates that the selected variables provide a stronger linear approximation to the fitted model response. Lower fidelity indicates that the coefficient columns should be read as a more limited selected-feature projection.

The second diagnostic measures coefficient stability under resampling. We bootstrap the explanation sample, refit the selected-feature surrogate on each bootstrap sample, and compare the resulting coefficient vectors with the original estimate. Let \(\hat\beta\) denote the original coefficient vector and \(\hat\beta^{(b)}\) the vector estimated from bootstrap sample \(b=1,\ldots,B\). The relative bootstrap deviation is
\[
    D^{(b)}
    =
    \frac{
        \|\hat\beta^{(b)}-\hat\beta\|_2
    }{
        \|\hat\beta\|_2+\epsilon
    },
\]
where \(\epsilon>0\) prevents division by zero. We summarize this variation by
\[
    \mathrm{BDisp}
    =
    \frac{1}{B}
    \sum_{b=1}^{B}
    D^{(b)}
\]
and
\[
    \mathrm{BStab}
    =
    \frac{1}{B}
    \sum_{b=1}^{B}
    \frac{1}{1+D^{(b)}} .
\]
Lower BDisp and higher BStab indicate a more stable coefficient vector. These diagnostics capture variation in coefficient magnitudes and joint coefficient structure, rather than only sign changes.

Together, fidelity and bootstrap stability define the scope of the table. Fidelity measures approximation quality, while bootstrap stability measures the reproducibility of the coefficient structure under resampling. The \(\phi\)-table should therefore be read as a post-hoc statistical summary of \(f(X)\) whose strength depends on both the selected-feature surrogate fit and the stability of its coefficients.

\section{Theory}
\label{sec:theory}

Throughout this section, \(f\) denotes a trained black-box model and
\[
Z=f(X)
\]
denotes its fitted response. For a selected feature set \(S\), the \(\phi\)-table summarizes \(Z\) through a standardized linear projection onto \(X_S\). Detailed proofs are given in Appendix~\ref{app:proofs}; the arguments below state the main objects and the key algebraic steps.

\subsection{Importance and Direction Are Non-Equivalent}
\label{subsec:importance_direction_nonequivalence}

Global SHAP importance is a magnitude summary. For feature \(j\),
\[
I_j=\mathbb{E}\{|\phi_j(X)|\},
\]
where \(\phi_j(X)\) is the SHAP attribution of feature \(j\). Since \(I_j\) is nonnegative and sign-invariant, direction enters through a different object: the projection coefficient of the fitted response on the selected standardized covariates.

\begin{proposition}[Non-equivalence of importance and direction]
\label{prop:importance_direction_nonequivalence}
There exists a distribution of \(X=(X_1,X_2)\) and a fitted response \(f\) such that
\[
I_1>I_2
\qquad\text{but}\qquad
|\beta_1^\star|<|\beta_2^\star|,
\]
where \(I_j=\mathbb{E}\{|\phi_j(X)|\}\) is the global SHAP importance and \(\beta_j^\star\) is the coefficient of feature \(j\) in the population linear projection of \(f(X)\) onto \((X_1,X_2)\). In particular, a feature may have positive global SHAP importance while having zero projection direction.
\end{proposition}

\begin{proof}
Let \(X_1\) and \(X_2\) be independent, mean-zero, unit-variance variables, with \(X_1\) symmetric, and define
\[
f(X_1,X_2)
=
a\{X_1^2-\mathbb{E}(X_1^2)\}+cX_2.
\]
For independent additive features, the SHAP attributions are the centered additive components:
\[
\phi_1(X)=a\{X_1^2-\mathbb{E}(X_1^2)\},
\qquad
\phi_2(X)=cX_2.
\]
Thus
\[
I_1
=
|a|\mathbb{E}\left|X_1^2-\mathbb{E}(X_1^2)\right|,
\qquad
I_2
=
|c|\mathbb{E}|X_2|.
\]
Choosing \(|a|\) sufficiently large relative to \(|c|\) gives \(I_1>I_2\). However,
\[
\mathbb{E}[X_1f(X_1,X_2)]
=
a\mathbb{E}\left[X_1\{X_1^2-\mathbb{E}(X_1^2)\}\right]
+
c\mathbb{E}[X_1X_2]
=
0,
\]
while
\[
\mathbb{E}[X_2f(X_1,X_2)]=c.
\]
Hence \(\beta_1^\star=0\) and \(\beta_2^\star\neq 0\), proving the claim. Standardizing \(Z\) only rescales both projection coefficients by a positive constant.
\end{proof}

\subsection{The Model-Response Projection Target}
\label{subsec:model_response_projection_target}

For a selected feature set \(S\), define
\[
\tilde Z=\frac{Z-\mathbb{E}[Z]}{\sqrt{\operatorname{Var}(Z)}},
\qquad
\tilde X_S
=
D_S^{-1}(X_S-\mathbb{E}[X_S]),
\]
where \(D_S\) is the diagonal matrix of marginal standard deviations of \(X_S\). The population coefficient underlying the \(\phi\)-table is
\[
\beta_S^\star
=
\arg\min_{\beta\in\mathbb{R}^{|S|}}
\mathbb{E}
\left[
\left(
\tilde Z-\beta^\top \tilde X_S
\right)^2
\right].
\]

\begin{proposition}[Model-response projection target]
\label{prop:model_response_projection_target}
Suppose \(\mathbb{E}[\tilde Z^2]<\infty\), \(\mathbb{E}[\|\tilde X_S\|_2^2]<\infty\), and
\[
\Sigma_S
=
\mathbb{E}[\tilde X_S\tilde X_S^\top]
\]
is nonsingular. Then the population projection coefficient is unique and satisfies
\[
\beta_S^\star
=
\Sigma_S^{-1}
\mathbb{E}[\tilde X_S\tilde Z].
\]
Moreover, the projection residual
\[
r_S
=
\tilde Z-\beta_S^{\star\top}\tilde X_S
\]
satisfies
\[
\mathbb{E}[\tilde X_S r_S]=0.
\]
\end{proposition}

\begin{proof}
Let
\[
Q(\beta)
=
\mathbb{E}
\left[
\left(
\tilde Z-\beta^\top \tilde X_S
\right)^2
\right],
\qquad
g_S=\mathbb{E}[\tilde X_S\tilde Z].
\]
Expanding the quadratic objective gives
\[
Q(\beta)
=
\mathbb{E}[\tilde Z^2]
-
2\beta^\top g_S
+
\beta^\top\Sigma_S\beta.
\]
Since \(\Sigma_S\) is positive definite, \(Q\) is strictly convex. The first-order condition
\[
-2g_S+2\Sigma_S\beta=0
\]
gives
\[
\beta_S^\star=\Sigma_S^{-1}g_S.
\]
Substituting this coefficient into \(r_S\) yields
\[
\mathbb{E}[\tilde X_S r_S]
=
g_S-\Sigma_S\beta_S^\star
=
0.
\]
\end{proof}

\subsection{Fidelity as Projection Strength}
\label{subsec:fidelity_projection_strength}

Let
\[
p_S^\star(X)
=
\beta_S^{\star\top}\tilde X_S
\]
be the population projection of \(\tilde Z\) onto the selected feature coordinates, with residual
\[
r_S
=
\tilde Z-p_S^\star(X).
\]
The population fidelity is
\[
F_S^\star
=
1-
\frac{
\mathbb{E}[r_S^2]
}{
\mathbb{E}[\tilde Z^2]
}.
\]

\begin{proposition}[Fidelity as projection strength]
\label{prop:fidelity_projection_strength}
Under the conditions of Proposition~\ref{prop:model_response_projection_target},
\[
0\leq F_S^\star\leq 1.
\]
Moreover,
\[
F_S^\star
=
\frac{
\mathbb{E}\left[(p_S^\star(X))^2\right]
}{
\mathbb{E}[\tilde Z^2]
}.
\]
If \(\operatorname{Var}(p_S^\star(X))>0\), then
\[
F_S^\star
=
\operatorname{Corr}^2
\left(
\tilde Z,
p_S^\star(X)
\right).
\]
When \(\mathbb{E}[\tilde Z^2]=1\),
\[
F_S^\star
=
g_S^\top \Sigma_S^{-1}g_S,
\qquad
g_S=\mathbb{E}[\tilde X_S\tilde Z].
\]
\end{proposition}

\begin{proof}
By Proposition~\ref{prop:model_response_projection_target},
\[
\mathbb{E}[\tilde X_S r_S]=0.
\]
Since \(p_S^\star(X)=\beta_S^{\star\top}\tilde X_S\), this implies
\[
\mathbb{E}[p_S^\star(X)r_S]=0.
\]
The orthogonal decomposition
\[
\tilde Z=p_S^\star(X)+r_S
\]
therefore gives
\[
\mathbb{E}[\tilde Z^2]
=
\mathbb{E}\left[(p_S^\star(X))^2\right]
+
\mathbb{E}[r_S^2].
\]
Rearranging yields the fidelity representation and \(0\leq F_S^\star\leq 1\). If \(p_S^\star(X)\) has positive variance, then
\[
\operatorname{Cov}(\tilde Z,p_S^\star(X))
=
\mathbb{E}\left[(p_S^\star(X))^2\right],
\]
which gives the squared-correlation form. Finally,
\[
\mathbb{E}\left[(p_S^\star(X))^2\right]
=
\beta_S^{\star\top}\Sigma_S\beta_S^\star
=
g_S^\top\Sigma_S^{-1}g_S.
\]
\end{proof}

\subsection{Uncertainty for Projection Coefficients}
\label{subsec:projection_coefficient_uncertainty}

Let \(\{X_i\}_{i=1}^m\) be the explanation sample, with
\[
Z_i=f(X_i),
\qquad
\tilde Z_i=\frac{Z_i-\mathbb{E}[Z]}{\sqrt{\operatorname{Var}(Z)}},
\qquad
\tilde X_{i,S}=D_S^{-1}(X_{i,S}-\mathbb{E}[X_S]).
\]
For a fixed selected feature set \(S\),
\[
\hat\beta_S
=
\arg\min_{\beta\in\mathbb{R}^{|S|}}
\frac{1}{m}
\sum_{i=1}^m
\left(
\tilde Z_i-\beta^\top \tilde X_{i,S}
\right)^2.
\]
Let
\[
r_{i,S}
=
\tilde Z_i-\beta_S^{\star\top}\tilde X_{i,S}.
\]

\begin{proposition}[Asymptotic uncertainty of projection coefficients]
\label{prop:projection_coefficient_uncertainty}
Suppose the explanation sample is i.i.d., \(S\) is fixed with \(|S|<\infty\),
\[
\Sigma_S=\mathbb{E}[\tilde X_S\tilde X_S^\top]
\]
is nonsingular, and
\[
\mathbb{E}\|\tilde X_S\|_2^4<\infty,
\qquad
\mathbb{E}[r_S^4]<\infty.
\]
Then
\[
\hat\beta_S
\overset{p}{\longrightarrow}
\beta_S^\star,
\]
and
\[
\sqrt{m}
\left(
\hat\beta_S-\beta_S^\star
\right)
\overset{d}{\longrightarrow}
N
\left(
0,
\Sigma_S^{-1}\Omega_S\Sigma_S^{-1}
\right),
\]
where
\[
\Omega_S
=
\mathbb{E}
\left[
\tilde X_S\tilde X_S^\top r_S^2
\right].
\]
Consequently,
\[
\operatorname{se}(\hat\beta_j)
=
\sqrt{
\frac{
\left[
\Sigma_S^{-1}\Omega_S\Sigma_S^{-1}
\right]_{jj}
}{m}
}.
\]
\end{proposition}

\begin{proof}
Define
\[
\hat\Sigma_S
=
\frac{1}{m}
\sum_{i=1}^m
\tilde X_{i,S}\tilde X_{i,S}^\top,
\qquad
\hat g_S
=
\frac{1}{m}
\sum_{i=1}^m
\tilde X_{i,S}\tilde Z_i.
\]
Then
\[
\hat\beta_S=\hat\Sigma_S^{-1}\hat g_S.
\]
The law of large numbers gives
\[
\hat\Sigma_S\overset{p}{\longrightarrow}\Sigma_S,
\qquad
\hat g_S\overset{p}{\longrightarrow}g_S,
\]
and hence \(\hat\beta_S\overset{p}{\longrightarrow}\beta_S^\star\). Moreover,
\[
\hat\beta_S-\beta_S^\star
=
\hat\Sigma_S^{-1}
\left(
\frac{1}{m}
\sum_{i=1}^m
\tilde X_{i,S}r_{i,S}
\right).
\]
Since \(\mathbb{E}[\tilde X_S r_S]=0\), the multivariate central limit theorem gives
\[
\frac{1}{\sqrt{m}}
\sum_{i=1}^m
\tilde X_{i,S}r_{i,S}
\overset{d}{\longrightarrow}
N(0,\Omega_S).
\]
Slutsky's theorem yields the stated asymptotic distribution and standard-error expression.
\end{proof}

The covariance term \(\Omega_S\) is estimated from empirical projection residuals. With
\[
\hat r_i
=
\tilde Z_i-\hat\beta_S^\top\tilde X_{i,S},
\]
and leverage values
\[
h_i
=
\tilde X_{i,S}^\top
\left(
\sum_{\ell=1}^m
\tilde X_{\ell,S}\tilde X_{\ell,S}^\top
\right)^{-1}
\tilde X_{i,S},
\]
the HC3 covariance estimator is
\[
\widehat V_{\mathrm{HC3}}
=
\hat\Sigma_S^{-1}
\left[
\frac{1}{m}
\sum_{i=1}^m
\tilde X_{i,S}\tilde X_{i,S}^\top
\left(
\frac{\hat r_i}{1-h_i}
\right)^2
\right]
\hat\Sigma_S^{-1}.
\]
The corresponding standard error is
\[
\widehat{\operatorname{se}}_{\mathrm{HC3}}(\hat\beta_j)
=
\sqrt{
\frac{
[\widehat V_{\mathrm{HC3}}]_{jj}
}{m}
},
\]
and the reported interval is
\[
\hat\beta_j
\pm
q_{1-\gamma/2}
\widehat{\operatorname{se}}_{\mathrm{HC3}}(\hat\beta_j),
\]
where \(q_{1-\gamma/2}\) is a normal or finite-sample \(t\) critical value.

\subsection{Coefficient Stability under Feature-Space Perturbations}
\label{subsec:coefficient_stability_perturbation}

The population projection coefficient can be written as
\[
\beta_S^\star
=
\Sigma_S^{-1}g_S,
\qquad
\Sigma_S=\mathbb{E}[\tilde X_S\tilde X_S^\top],
\qquad
g_S=\mathbb{E}[\tilde X_S\tilde Z].
\]
Let
\[
\Sigma_S^{\Delta}
=
\Sigma_S+\Delta_{\Sigma},
\qquad
g_S^{\Delta}
=
g_S+\Delta_g,
\]
and define
\[
\beta_S^{\Delta}
=
(\Sigma_S+\Delta_{\Sigma})^{-1}
(g_S+\Delta_g).
\]

\begin{proposition}[Coefficient sensitivity under feature-space perturbations]
\label{prop:coefficient_stability_perturbation}
Suppose \(\Sigma_S\) is nonsingular and
\[
\left\|
\Sigma_S^{-1}\Delta_{\Sigma}
\right\|_{\mathrm{op}}
<1.
\]
Then \(\Sigma_S+\Delta_{\Sigma}\) is nonsingular and
\[
\beta_S^{\Delta}-\beta_S^\star
=
(\Sigma_S+\Delta_{\Sigma})^{-1}
\left(
\Delta_g-\Delta_{\Sigma}\beta_S^\star
\right).
\]
Consequently,
\[
\left\|
\beta_S^{\Delta}-\beta_S^\star
\right\|_2
\leq
\frac{
\|\Sigma_S^{-1}\|_{\mathrm{op}}
}{
1-\|\Sigma_S^{-1}\Delta_{\Sigma}\|_{\mathrm{op}}
}
\left(
\|\Delta_g\|_2
+
\|\Delta_{\Sigma}\|_{\mathrm{op}}
\|\beta_S^\star\|_2
\right).
\]
\end{proposition}

\begin{proof}
The condition
\[
\|\Sigma_S^{-1}\Delta_{\Sigma}\|_{\mathrm{op}}<1
\]
implies that
\[
\Sigma_S+\Delta_{\Sigma}
=
\Sigma_S(I+\Sigma_S^{-1}\Delta_{\Sigma})
\]
is nonsingular. Since \(g_S=\Sigma_S\beta_S^\star\),
\[
\beta_S^\Delta-\beta_S^\star
=
(\Sigma_S+\Delta_{\Sigma})^{-1}
\left(
\Delta_g-\Delta_{\Sigma}\beta_S^\star
\right).
\]
Taking norms gives
\[
\left\|
\beta_S^\Delta-\beta_S^\star
\right\|_2
\leq
\left\|
(\Sigma_S+\Delta_{\Sigma})^{-1}
\right\|_{\mathrm{op}}
\left(
\|\Delta_g\|_2
+
\|\Delta_{\Sigma}\|_{\mathrm{op}}
\|\beta_S^\star\|_2
\right).
\]
The inverse satisfies
\[
(\Sigma_S+\Delta_{\Sigma})^{-1}
=
(I+\Sigma_S^{-1}\Delta_{\Sigma})^{-1}\Sigma_S^{-1},
\]
and the Neumann-series bound gives
\[
\left\|
(\Sigma_S+\Delta_{\Sigma})^{-1}
\right\|_{\mathrm{op}}
\leq
\frac{
\|\Sigma_S^{-1}\|_{\mathrm{op}}
}{
1-\|\Sigma_S^{-1}\Delta_{\Sigma}\|_{\mathrm{op}}
}.
\]
\end{proof}

For the empirical diagnostic, let \(\hat\beta_S\) be the coefficient vector from the original explanation sample and \(\hat\beta_S^{(b)}\) the coefficient vector from bootstrap resample \(b\). The relative bootstrap deviation is
\[
D^{(b)}
=
\frac{
\|\hat\beta_S^{(b)}-\hat\beta_S\|_2
}{
\|\hat\beta_S\|_2+\epsilon
},
\]
with \(\epsilon>0\). The reported summaries are
\[
BDisp
=
\frac{1}{B}
\sum_{b=1}^{B}
D^{(b)},
\qquad
BStab
=
\frac{1}{B}
\sum_{b=1}^{B}
\frac{1}{1+D^{(b)}}.
\]

\subsection{Fidelity and Stability Are Non-Substitutable}
\label{subsec:fidelity_stability_nonsubstitutable}

Fidelity concerns the amount of fitted-response variation captured by the selected linear projection. Stability concerns the robustness of the coefficient allocation within that projection. The two quantities can separate sharply.

\begin{proposition}[Fidelity and stability are non-substitutable]
\label{prop:fidelity_stability_nonsubstitutable}
The following two statements hold.

\begin{enumerate}
    \item For any \(M>0\), there exists a fitted response \(f\) and a selected feature set \(S\) such that
    \[
    F_S^\star=1,
    \]
    but the projection coefficient can be amplified by more than \(M\) under an arbitrarily small perturbation of \(g_S\).

    \item For any \(\eta>0\), there exists a fitted response \(f\) and a selected feature set \(S\) such that the projection coefficient is well-conditioned, but
    \[
    F_S^\star<\eta.
    \]
\end{enumerate}
\end{proposition}

\begin{proof}
For the first statement, let \(S=\{1,2\}\), and let \((X_1,X_2)\) be standardized with
\[
\Sigma_S
=
\begin{pmatrix}
1 & \rho \\
\rho & 1
\end{pmatrix},
\qquad
0<\rho<1.
\]
Set
\[
Z=f(X)=X_1+X_2.
\]
Then \(Z\) lies in the span of \(X_S\), so \(F_S^\star=1\). Perturb \(g_S\) by
\[
\Delta_g
=
\delta
\begin{pmatrix}
1\\
-1
\end{pmatrix}.
\]
Since
\[
\Sigma_S^{-1}
\begin{pmatrix}
1\\
-1
\end{pmatrix}
=
\frac{1}{1-\rho}
\begin{pmatrix}
1\\
-1
\end{pmatrix},
\]
the perturbation is amplified by the factor \((1-\rho)^{-1}\), which can exceed any \(M\) by taking \(\rho\) close to one.

For the second statement, let \(S=\{1\}\), let \(X\) be standardized and symmetric, and set
\[
Z=f(X)
=
aX+b\{X^2-\mathbb{E}(X^2)\},
\qquad
a\neq 0.
\]
The linear and quadratic components are orthogonal, while \(\Sigma_S=1\). Hence the coefficient map is well-conditioned and
\[
F_S^\star
=
\frac{
a^2
}{
a^2+b^2\operatorname{Var}(X^2)
}.
\]
Choosing \(|b|\) sufficiently large makes \(F_S^\star<\eta\).
\end{proof}

\section{Experiments}
\label{sec:experiments}

\subsection{Common Experimental Design}
\label{subsec:common_exp_design}

\paragraph{Datasets and models.}
The experiments combine one controlled synthetic study with three real-data studies. Experiment~1 uses a synthetic response function designed to separate feature importance from directional effects. Experiments~2--4 are conducted on seven tabular regression datasets: Air Quality, Airfoil Self-Noise, Bike Sharing, California Housing, Combined Cycle Power Plant, Concrete Compressive Strength, and Energy Efficiency. For these datasets, we retain usable numeric covariates, remove columns with excessive missingness, impute remaining missing values using the median, and standardize the covariates. When the original response is used, the target is standardized as well. A random forest regressor serves as the black-box model in Experiments~2--4.

\paragraph{Experimental settings.}
Experiment~2 uses a semi-synthetic response: the covariate distribution comes from real datasets, but the response is generated from a known function with linear, nonlinear, sinusoidal, and interaction components. This design provides oracle projection coefficients against which the estimated table coefficients can be compared. Experiments~3 and~4 instead use the original dataset targets to train the black-box model, so their role is to evaluate table diagnostics under ordinary tabular regression conditions.

\paragraph{Evaluation metrics.}
The metrics are grouped by the type of information they evaluate:
\begin{itemize}
    \item \textbf{Importance--coefficient alignment.}
    RankCorr measures the rank correlation between SHAP importance and absolute surrogate coefficient magnitude. Disc@k measures the fraction of SHAP top-$k$ features that are not also selected by coefficient-based criteria.

    \item \textbf{Directional clarity.}
    DirRate measures the fraction of SHAP-selected features whose confidence interval excludes zero, while Ambig@k measures the fraction whose confidence interval includes zero. NonDir@k measures the fraction of selected features without clear directional summary in the controlled synthetic setting. SignAcc measures whether surrogate coefficients recover the expected or oracle coefficient sign.

    \item \textbf{False directional assignment.}
    FalseDir measures how often the surrogate assigns a strong directional interpretation to features that are non-directional by construction, such as symmetric nonlinear, interaction-only, or noise features.

    \item \textbf{Projection recovery.}
    BetaErr measures the absolute error between estimated surrogate coefficients and oracle projection coefficients. CICov measures how often the confidence interval contains the oracle projection coefficient.

    \item \textbf{Coefficient uncertainty.}
    CIW summarizes coefficient-level uncertainty by averaging the confidence-interval widths of the selected features.

    \item \textbf{Surrogate fidelity.}
    Fid measures how well the selected-feature surrogate explains the fitted black-box response $f(X)$. In Experiment~2, this is evaluated as held-out surrogate fidelity; in Experiments~3 and~4, it is used as descriptive fidelity on the explanation sample.

    \item \textbf{Bootstrap coefficient stability.}
    BStab measures coefficient-vector stability under bootstrap resampling of the explanation sample. BDisp measures the corresponding coefficient-vector dispersion. FidCorr measures the within-dataset correlation between fidelity and bootstrap coefficient stability, and FidGap compares coefficient stability between high-fidelity and low-fidelity repetitions.
\end{itemize}

\subsection{Experiment 1: Importance--Direction Dissociation}
\label{subsec:exp1}

\paragraph{Experimental setup.}
To isolate the distinction between importance and direction, we construct a synthetic response with known feature roles. The data contain two directional linear features, a symmetric nonlinear feature, an interaction pair, a correlated proxy, a redundant mixture feature, and a noise feature. Because the response function is known, exact SHAP values can be computed directly. We then rank features by mean absolute SHAP value, select the top-$k$ features, and estimate a standardized linear surrogate for the black-box response with HC3 confidence intervals. The experiment uses 100 independent replications, each with 2,500 observations and an explanation sample of 1,200 observations.

\paragraph{Results and analysis.}
Table~\ref{tab:exp1_importance_direction} shows a clear separation between SHAP importance and surrogate coefficient direction. Although the signs of the truly directional linear features are recovered without error, many SHAP-selected features are classified as non-directional, and the rank correlation between SHAP importance and absolute coefficient magnitude remains only moderate. This pattern is expected in the controlled design, where proxy, redundant, and interaction-related variables can receive high attribution while admitting weak or unstable linear directional summaries. The low false-direction rate further indicates that the $\phi$-table does not simply force every important feature into a positive or negative interpretation; instead, most top-$k$ discordance is concentrated among selected features with limited directional evidence.

\begin{table}[ht]
\centering
\caption{Results for Experiment 1. The table summarizes the separation between SHAP importance and surrogate coefficient direction over 100 replications.}
\label{tab:exp1_importance_direction}
\begin{tabular}{lccccc}
\hline
Summary & RankCorr & Disc@k & NonDir@k & SignAcc & FalseDir \\
\hline
Estimate & 0.464 $\pm$ 0.258 & 0.558 $\pm$ 0.091 & 0.556 $\pm$ 0.092 & 1.000 $\pm$ 0.000 & 0.073 $\pm$ 0.154 \\
\hline
\end{tabular}
\end{table}

\subsection{Experiment 2: Projection Recovery and Uncertainty Coverage}
\label{subsec:exp2}

\paragraph{Experimental setup.}
Projection validity is evaluated with a semi-synthetic design. For each real tabular dataset, we keep the empirical covariate distribution but replace the original target with a generated response containing linear, nonlinear, sinusoidal, and interaction components. A random forest is trained on this generated response, after which SHAP values are computed on the explanation split and the top-$k$ features are selected. The estimated table coefficients are obtained by fitting the selected-feature surrogate on the explanation split. As a reference target, we compute the corresponding full-sample projection coefficients of the standardized black-box output onto the same selected features. Each dataset is evaluated over 30 replications.

\paragraph{Results and analysis.}
Table~\ref{tab:exp2_projection_validity} shows that the selected-feature surrogate often recovers the model-response projection accurately, but that recovery depends on how well the fitted response can be represented by a global linear summary. Airfoil Self-Noise, Concrete Compressive Strength, and Air Quality exhibit the strongest recovery patterns, combining small coefficient error, high sign accuracy, strong interval coverage, and high fidelity. Bike Sharing and Energy Efficiency preserve reliable coefficient signs and coverage even when fidelity is lower, while Combined Cycle Power Plant achieves small coefficient error despite a weaker linear approximation. California Housing serves as the main stress case: its larger coefficient error, weaker sign recovery, lower coverage, and greater fidelity variation indicate that the fitted response is harder to summarize through a single selected-feature linear projection. This stress case illustrates the role of the fidelity diagnostic: coefficient recovery and directional interpretation should be read together with how well the selected-feature surrogate summarizes the fitted response.

\begin{table}[ht]
\centering
\caption{Results for Experiment 2. The table evaluates recovery of model-response projection coefficients and held-out surrogate fidelity over 30 replications.}
\label{tab:exp2_projection_validity}
\begin{tabular}{lcccc}
\hline
Dataset & BetaErr & SignAcc & CICov & Fid \\
\hline
Air Quality & 0.098 $\pm$ 0.055 & 1.000 $\pm$ 0.000 & 0.973 $\pm$ 0.087 & 0.859 $\pm$ 0.029 \\
Airfoil Self-Noise & 0.019 $\pm$ 0.006 & 1.000 $\pm$ 0.000 & 0.987 $\pm$ 0.051 & 0.861 $\pm$ 0.011 \\
Bike Sharing & 0.012 $\pm$ 0.004 & 0.907 $\pm$ 0.101 & 0.973 $\pm$ 0.069 & 0.715 $\pm$ 0.013 \\
California Housing & 0.177 $\pm$ 0.113 & 0.747 $\pm$ 0.157 & 0.747 $\pm$ 0.246 & 0.221 $\pm$ 0.959 \\
Combined Cycle Power Plant & 0.034 $\pm$ 0.015 & 0.967 $\pm$ 0.086 & 0.992 $\pm$ 0.046 & 0.067 $\pm$ 0.015 \\
Concrete Compressive Strength & 0.024 $\pm$ 0.010 & 1.000 $\pm$ 0.000 & 0.987 $\pm$ 0.051 & 0.854 $\pm$ 0.020 \\
Energy Efficiency & 0.134 $\pm$ 0.068 & 0.927 $\pm$ 0.098 & 1.000 $\pm$ 0.000 & 0.633 $\pm$ 0.040 \\
\hline
\end{tabular}
\end{table}

\subsection{Experiment 3: Information Gain over Ranking-Only Explanation}
\label{subsec:exp3}

\paragraph{Experimental setup.}
This experiment uses the original targets of the seven tabular datasets to examine what the table adds beyond a SHAP importance ranking. After training the random forest on each dataset, we explain the fitted response $f(X)$ rather than the observed outcome. SHAP values determine the top-$k$ feature set, and the selected features are used to fit a linear surrogate with HC3 confidence intervals. The resulting summaries record whether SHAP-selected features have clear coefficient directions, whether they are ambiguous, how much the SHAP top-$k$ set differs from the coefficient-based top-$k$ set, how wide the coefficient intervals are, and how well the selected-feature surrogate fits the black-box response. Results are aggregated over 20 replications per dataset.

\paragraph{Results and analysis.}
Table~\ref{tab:exp3_information_gain} shows that the additional $\phi$-table columns provide information that is not contained in the SHAP ranking alone. Airfoil Self-Noise and Combined Cycle Power Plant display full directional clarity and no top-$k$ discordance, indicating close agreement between attribution-based selection and coefficient-based summaries. By contrast, Bike Sharing shows high ambiguity, large discordance, and low fidelity, suggesting that important features do not necessarily yield stable global directions. Air Quality and California Housing also show meaningful disagreement between SHAP selection and coefficient-based ranking, with California Housing further marked by wide confidence intervals. These patterns show that direction, ambiguity, uncertainty, discordance, and fidelity are distinct components of explanation quality rather than redundant restatements of SHAP importance.

\begin{table}[ht]
\centering
\caption{Results for Experiment 3. The table summarizes the additional diagnostic information provided by the statistical explanation table beyond ranking-only SHAP over 20 replications.}
\label{tab:exp3_information_gain}
\resizebox{\textwidth}{!}{
\begin{tabular}{lccccc}
\hline
Dataset & DirRate & Ambig@k & Disc@k & CIW & Fid \\
\hline
Air Quality & 0.810 $\pm$ 0.152 & 0.190 $\pm$ 0.152 & 0.300 $\pm$ 0.103 & 0.133 $\pm$ 0.018 & 0.959 $\pm$ 0.005 \\
Airfoil Self-Noise & 1.000 $\pm$ 0.000 & 0.000 $\pm$ 0.000 & 0.000 $\pm$ 0.000 & 0.216 $\pm$ 0.016 & 0.635 $\pm$ 0.032 \\
Bike Sharing & 0.640 $\pm$ 0.154 & 0.360 $\pm$ 0.154 & 0.340 $\pm$ 0.114 & 0.163 $\pm$ 0.094 & 0.398 $\pm$ 0.022 \\
California Housing & 0.880 $\pm$ 0.120 & 0.120 $\pm$ 0.120 & 0.290 $\pm$ 0.121 & 1.714 $\pm$ 1.293 & 0.796 $\pm$ 0.024 \\
Combined Cycle Power Plant & 1.000 $\pm$ 0.000 & 0.000 $\pm$ 0.000 & 0.000 $\pm$ 0.000 & 0.034 $\pm$ 0.001 & 0.976 $\pm$ 0.001 \\
Concrete Compressive Strength & 0.980 $\pm$ 0.062 & 0.020 $\pm$ 0.062 & 0.160 $\pm$ 0.105 & 0.247 $\pm$ 0.025 & 0.611 $\pm$ 0.042 \\
Energy Efficiency & 0.880 $\pm$ 0.151 & 0.120 $\pm$ 0.151 & 0.040 $\pm$ 0.082 & 0.373 $\pm$ 0.078 & 0.930 $\pm$ 0.010 \\
\hline
\end{tabular}
}
\end{table}

\subsection{Experiment 4: Fidelity- and Stability-Conditioned Table Interpretation}
\label{subsec:exp4}

\paragraph{Experimental setup.}
The final experiment measures how stable the table coefficients are under resampling. Using the same real-data protocol as Experiment~3, we train the black-box model, select the SHAP top-$k$ features, and fit the selected-feature surrogate to the standardized fitted response. Within each replication, the explanation sample is then bootstrapped 50 times. For every bootstrap sample, the surrogate is refit and its coefficient vector is compared with the original coefficient vector through relative deviation. This yields a bootstrap stability score, a dispersion measure, and two summaries of the relationship between fidelity and stability across repeated splits. Each dataset is evaluated over 20 replications.

\paragraph{Results and analysis.}
Table~\ref{tab:exp4_fidelity_stability} shows that surrogate fidelity and coefficient stability capture related but distinct aspects of the $\phi$-table. Air Quality and Combined Cycle Power Plant combine high fidelity with high bootstrap stability, while Airfoil Self-Noise and Concrete Compressive Strength maintain stable coefficient vectors despite only moderate fidelity. Bike Sharing has the weakest linear approximation, yet its coefficient stability remains comparatively high on average, albeit with greater variability. California Housing is especially informative because moderate fidelity does not translate into stable coefficients: it has the weakest stability and the largest coefficient dispersion. Energy Efficiency shows the opposite kind of separation, combining high fidelity and high average stability with a slightly negative fidelity--stability correlation. Together, these results support reporting fidelity and stability side by side, since approximation quality alone does not determine whether the directional coefficient structure is reproducible.

\begin{table}[ht]
\centering
\caption{Results for Experiment 4. The table evaluates surrogate fidelity and bootstrap coefficient stability over 20 replications.}
\label{tab:exp4_fidelity_stability}
\begin{tabular}{lccccc}
\hline
Dataset & Fid & BStab & BDisp & FidCorr & FidGap \\
\hline
Air Quality & 0.957 $\pm$ 0.003 & 0.933 $\pm$ 0.011 & 0.072 $\pm$ 0.012 & 0.162 & 0.003 \\
Airfoil Self-Noise & 0.625 $\pm$ 0.020 & 0.924 $\pm$ 0.005 & 0.082 $\pm$ 0.005 & 0.496 & 0.006 \\
Bike Sharing & 0.400 $\pm$ 0.009 & 0.887 $\pm$ 0.069 & 0.134 $\pm$ 0.091 & 0.097 & 0.013 \\
California Housing & 0.777 $\pm$ 0.018 & 0.641 $\pm$ 0.151 & 0.651 $\pm$ 0.424 & 0.651 & 0.200 \\
Combined Cycle Power Plant & 0.976 $\pm$ 0.001 & 0.988 $\pm$ 0.001 & 0.012 $\pm$ 0.001 & 0.235 & 0.000 \\
Concrete Compressive Strength & 0.596 $\pm$ 0.017 & 0.892 $\pm$ 0.007 & 0.121 $\pm$ 0.009 & 0.553 & 0.010 \\
Energy Efficiency & 0.928 $\pm$ 0.005 & 0.890 $\pm$ 0.012 & 0.124 $\pm$ 0.015 & -0.118 & -0.005 \\
\hline
\end{tabular}
\end{table}

\section{Discussion}
\label{sec:discussion}

\subsection{From Ranking to Statistical Explanation}
\label{subsec:ranking_to_table}

Global SHAP is typically presented as an ordered list of features by attribution magnitude. This format is useful, but it compresses global explanation into a single question: which variables matter most to the fitted model response? The \(\phi\)-table keeps this screening role for SHAP while changing what is reported after selection. Once the top-ranked features are fixed, the table augments the ranking with model-response coefficients, uncertainty intervals, coefficient-level uncertainty summaries, surrogate fidelity, and coefficient stability. The resulting object is no longer only an importance list; it is a statistical summary of how the selected features relate to the black-box response.

This change in format also changes the interpretation. A high SHAP rank does not by itself say whether a feature admits a clear positive or negative global summary, whether that summary is stable, or whether the selected variables form a faithful low-dimensional approximation to the model. The \(\phi\)-table makes these additional conditions visible. It tells the reader not only which variables were selected, but also how strongly their directional summaries are supported and how broadly the table can be read as a summary of the fitted model.

\subsection{Reading the \texorpdfstring{$\phi$}{phi}-Table in Practice}
\label{subsec:reading_phi_table}

\begin{table}[ht]
\centering
\small
\setlength{\tabcolsep}{4pt}
\caption{Example \(\phi\)-table for Energy Efficiency with heating load as the target. The explanation target is the standardized random-forest output \(f(X)\); the black-box \(R^2\) on the explanation split is 0.987 and the surrogate fidelity \(R^2\) is 0.926.}
\label{tab:phi_table_energy}
\resizebox{\textwidth}{!}{
\begin{tabular}{rlrrrcclr}
\toprule
Rank & Feature & SHAP Imp. & Coef. & Std. Err. & 95\% Cond. CI & Cond. \(p\)-value & Direction & Coef. Stab. \\
\midrule
1 & Roof\_Area & 0.266 & -0.481 & 0.087 & [-0.651, -0.310] & \(<0.001\) & Negative & 0.881 \\
2 & Overall\_Height & 0.255 & 0.783 & 0.095 & [\hphantom{-}0.597, \hphantom{-}0.968] & \(<0.001\) & Positive & 0.918 \\
3 & Surface\_Area & 0.253 & -0.270 & 0.083 & [-0.432, -0.108] & 0.001 & Negative & 0.819 \\
4 & Glazing\_Area & 0.205 & 0.244 & 0.018 & [\hphantom{-}0.209, \hphantom{-}0.278] & \(<0.001\) & Positive & 0.949 \\
5 & Relative\_Compactness & 0.140 & -0.716 & 0.082 & [-0.878, -0.555] & \(<0.001\) & Negative & 0.922 \\
\bottomrule
\end{tabular}
}
\end{table}

\begin{table}[ht]
\centering
\small
\setlength{\tabcolsep}{4pt}
\caption{Example \(\phi\)-table for California Housing with median house value as the target. The explanation target is the standardized random-forest output \(f(X)\); the black-box \(R^2\) on the explanation split is 0.748 and the surrogate fidelity \(R^2\) is 0.765.}
\label{tab:phi_table_california}
\begin{tabular}{rlrrrcclr}
\toprule
Rank & Feature & SHAP Imp. & Coef. & Std. Err. & 95\% Cond. CI & Cond. \(p\)-value & Direction & Coef. Stab. \\
\midrule
1 & MedInc & 0.534 & 0.780 & 0.010 & [\hphantom{-}0.761, \hphantom{-}0.799] & \(<0.001\) & Positive & 0.990 \\
2 & AveOccup & 0.194 & -0.028 & 0.048 & [-0.122, \hphantom{-}0.066] & 0.558 & Ambiguous & 0.149 \\
3 & Latitude & 0.141 & -0.609 & 0.016 & [-0.640, -0.578] & \(<0.001\) & Negative & 0.980 \\
4 & Longitude & 0.099 & -0.574 & 0.016 & [-0.606, -0.541] & \(<0.001\) & Negative & 0.979 \\
5 & HouseAge & 0.081 & 0.146 & 0.006 & [\hphantom{-}0.133, \hphantom{-}0.158] & \(<0.001\) & Positive & 0.968 \\
\bottomrule
\end{tabular}
\end{table}

Tables~\ref{tab:phi_table_energy} and~\ref{tab:phi_table_california} show the kind of explanation produced by the method. In the Energy Efficiency example, the selected features form a high-fidelity surrogate and all coefficient intervals exclude zero. The table therefore supports a clear directional reading within the selected-feature surrogate: overall height and glazing area are associated with larger fitted heating-load predictions, whereas roof area, surface area, and relative compactness are associated with smaller predictions. The stability values reinforce this interpretation by showing that the coefficient summaries are reproducible under resampling.

The California Housing example shows why the additional columns are useful even when the SHAP ranking itself is informative. Median income is both the highest-ranked feature and a stable positive coefficient. Latitude and longitude have lower SHAP importance than median income, but their coefficients are large, negative, and stable. Average occupancy gives a different signal: it is ranked second by SHAP importance, yet its coefficient interval includes zero and its coefficient stability is weak. This contrast makes California Housing a useful stress case for ranking-only explanation. A feature can matter to the black-box model while failing to support a stable global directional summary in the selected-feature surrogate. Table~\ref{tab:phi_table_california} therefore illustrates the main value of the additional columns: they separate attribution magnitude from direction, uncertainty, and coefficient stability within the same global explanation.

\subsection{Importance, Direction, and Ambiguity}
\label{subsec:importance_direction_ambiguity}

The results reinforce that attribution magnitude and directional projection are different statistical objects. SHAP importance measures how strongly a feature contributes to the fitted response in absolute attribution terms, whereas the coefficient column summarizes the direction of a selected-feature projection. Nonlinear effects, symmetric responses, interactions, thresholds, and correlated proxies can therefore produce high importance without a stable positive or negative coefficient. The synthetic experiment isolates this separation by design, and the real-data examples show the same pattern in applied settings: some highly ranked features support clear coefficients, while others remain weak, unstable, or ambiguous. The value of the \(\phi\)-table is that it keeps these cases separated rather than translating every important feature into a directional claim.

Ambiguity is therefore not a failure of the table but one of its explanation signals. In a ranking-only display, a highly ranked feature can be read too quickly as having a clear direction in the model. The \(\phi\)-table makes that interpretation conditional on coefficient magnitude, interval width, sign clarity, and resampling stability. When a selected feature has an interval crossing zero or weak coefficient stability, the table indicates that the feature may matter to the black-box model without admitting a simple global directional summary. This is especially useful for tabular black-box models, where fitted responses often depend on interactions, nonlinear transformations, and region-specific behavior that a single global coefficient should not be forced to summarize.

\subsection{Projection Validity, Fidelity, and Stability}
\label{subsec:projection_fidelity_stability}

The coefficient columns have a well-defined target: they estimate the linear projection of the fitted model response onto the SHAP-selected covariates. This target concerns \(f(X)\), not the original outcome-generating process. The semi-synthetic experiment supports this interpretation by showing that the estimated coefficients can recover the corresponding model-response projection targets and that the confidence intervals provide useful uncertainty summaries for those targets. Projection validity, however, does not by itself determine the strength of the explanation. A coefficient may accurately estimate its projection target even when the selected-feature surrogate captures only a limited part of the black-box response.

Fidelity and stability therefore define the scope of interpretation. Fidelity measures how much of the fitted response is captured by the selected linear surrogate, while bootstrap stability measures how reproducible the coefficient vector is under resampling. These diagnostics need not move together: a high-fidelity surrogate can have unstable coefficients when selected covariates are correlated, and a stable coefficient vector can arise from a surrogate that captures only a restricted component of the fitted response. The strongest directional reading is supported when fidelity and stability are both high. Low fidelity narrows the table to a limited projection summary, and low stability weakens coefficient-level interpretation even when coefficient signs appear clear.

\subsection{Limitations and Future Work}
\label{subsec:limitations_future_work}

The \(\phi\)-table is a post-hoc summary of fitted model behavior through a selected-feature surrogate for \(f(X)\). Its coefficients, intervals, and \(p\)-values are interpreted within this surrogate after SHAP-based feature selection. A fuller treatment of the selection step would require selective-inference or resampling procedures that account for the ranking and top-\(k\) selection process itself. The present evaluation also focuses on tabular regression settings, where standardized fitted responses and regression-style summaries have a direct interpretation. Other prediction tasks may require different response scales, uncertainty summaries, and interpretation conventions.

Several extensions follow naturally from this formulation. For classification, the same framework could be applied to predicted probabilities, logits, or class scores, with the target scale chosen according to the explanation goal. Future work could also study alternative attribution methods, selection rules, black-box backbones, and uncertainty estimators, including bootstrap and selective-inference adjustments. When ambiguity or low fidelity indicates that a single global projection is too coarse, the table can be refined by increasing the selected feature set, adding nonlinear or interaction terms to the surrogate, or constructing subgroup-specific and local-region \(\phi\)-tables.

\section{Conclusion}
\label{sec:conclusion}

We proposed the $\phi$-table, a SHAP-based statistical explanation table that extends global feature-importance rankings with direction, uncertainty, surrogate fidelity, and coefficient stability. The table preserves SHAP as a model-aware screening mechanism while interpreting the resulting coefficients as projections of the fitted black-box response \(f(X)\) onto the SHAP-selected feature set. Across synthetic, semi-synthetic, and real tabular regression experiments, the results show that importance and direction can diverge, that the proposed coefficients recover model-response projection targets, and that fidelity and stability are necessary conditions for reading the table as a reliable global summary. These findings suggest that ranking-only explanations can be made more informative by pairing attribution importance with statistical summaries that clarify how strongly, how directionally, and how stably selected features describe trained model behavior.

\section*{Broader Impact Statement}

This work develops a statistical explanation format for trained tabular black-box regression models rather than a predictive model intended for direct deployment. Its main benefit is to make global explanations more informative and less prone to ranking-based overinterpretation. By augmenting SHAP importance with coefficient direction, uncertainty intervals, uncertainty summaries, surrogate fidelity, and bootstrap stability, the $\phi$-table helps analysts separate features that strongly contribute to model predictions from features that also support stable directional summaries. In this sense, the method can improve communication among model developers, auditors, and domain experts by making the scope and reliability of global model-response interpretation explicit.

The format must nevertheless be read within its intended target. The coefficients, confidence intervals, and p-values in the $\phi$-table summarize the fitted black-box response $f(X)$ through a selected-feature surrogate; they do not estimate causal effects, policy effects, or the data-generating mechanism behind the real outcome $Y$. This distinction is especially important in healthcare, lending, employment, insurance, education, public administration, and other high-stakes settings, where regression-like outputs can easily acquire unwarranted authority. The table should therefore support model diagnosis and review, not serve as a stand-alone basis for automated decisions, adverse actions, or policy interventions.

The proposed diagnostics are designed to expose these boundaries. Low surrogate fidelity indicates that the selected-feature table captures only a limited component of the black-box response. Low coefficient stability shows that directional summaries are sensitive to the explanation sample. Wide intervals or ambiguous coefficients signal that an important feature should not be compressed into a simple positive or negative global interpretation. These warnings become more consequential when models are trained on biased, incomplete, or historically inequitable data, since explanation tables can otherwise make problematic model behavior appear statistically settled.

The empirical evaluation uses standard tabular regression benchmarks together with synthetic and semi-synthetic designs, and does not involve human-subject data collection, direct user intervention, or deployment in operational systems. Future applications of the $\phi$-table should be paired with domain-specific validation, fairness auditing, data quality assessment, and human review. The broader aim is to provide a disciplined explanation table that makes uncertainty, fidelity, and stability visible whenever global directional interpretations of black-box behavior are reported.

\section*{Reproducibility}

All experiments reported in the main paper can be reproduced using a single notebook, \texttt{phi\_table.ipynb}, provided in the Supplementary Material. This file includes the data-loading, preprocessing, model-training, SHAP computation, $\phi$-table construction, and evaluation routines used throughout the paper.

\bibliography{main}
\bibliographystyle{tmlr}

\appendix

\section{Proofs}
\label{app:proofs}

\subsection{Proof of Proposition~\ref{prop:importance_direction_nonequivalence}}
\label{app:proof_importance_direction_nonequivalence}

\begin{proof}
Let \(X_1\) and \(X_2\) be independent, mean-zero, unit-variance random variables. Assume that \(X_1\) is symmetric, so that
\[
\mathbb{E}[X_1^3]=0.
\]
Consider the additive fitted response
\[
f(X_1,X_2)
=
a\{X_1^2-\mathbb{E}(X_1^2)\}+cX_2,
\]
where \(a\neq 0\) and \(c\neq 0\). Since the model is additive and the features are independent, the SHAP attribution of each feature is its centered additive component:
\[
\phi_1(X)
=
a\{X_1^2-\mathbb{E}(X_1^2)\},
\qquad
\phi_2(X)
=
cX_2.
\]
Therefore the global SHAP importances are
\[
I_1
=
|a|\,
\mathbb{E}
\left[
\left|
X_1^2-\mathbb{E}(X_1^2)
\right|
\right],
\qquad
I_2
=
|c|\,
\mathbb{E}[|X_2|].
\]
Because
\[
\mathbb{E}
\left[
\left|
X_1^2-\mathbb{E}(X_1^2)
\right|
\right]
>0
\]
whenever \(X_1^2\) is nondegenerate, we can choose \(|a|/|c|\) sufficiently large so that
\[
I_1>I_2.
\]

Now consider the population linear projection of \(f(X)\) onto \((X_1,X_2)\). Since \(X_1\) and \(X_2\) are independent and have unit variance, the projection coefficient for \(X_j\) is determined by \(\mathbb{E}[X_j f(X)]\). For \(X_1\),
\[
\begin{aligned}
\mathbb{E}[X_1 f(X_1,X_2)]
&=
a\,
\mathbb{E}
\left[
X_1
\{X_1^2-\mathbb{E}(X_1^2)\}
\right]
+
c\,\mathbb{E}[X_1X_2] \\
&=
a
\left\{
\mathbb{E}[X_1^3]
-
\mathbb{E}(X_1^2)\mathbb{E}[X_1]
\right\}
+
c\,\mathbb{E}[X_1]\mathbb{E}[X_2] \\
&=0.
\end{aligned}
\]
Thus
\[
\beta_1^\star=0.
\]
For \(X_2\),
\[
\begin{aligned}
\mathbb{E}[X_2 f(X_1,X_2)]
&=
a\,
\mathbb{E}
\left[
X_2
\{X_1^2-\mathbb{E}(X_1^2)\}
\right]
+
c\,\mathbb{E}[X_2^2] \\
&=
a\,
\mathbb{E}[X_2]\,
\mathbb{E}
\left[
X_1^2-\mathbb{E}(X_1^2)
\right]
+
c \\
&=c.
\end{aligned}
\]
Since \(c\neq 0\),
\[
\beta_2^\star\neq 0.
\]
Therefore,
\[
I_1>I_2
\qquad\text{but}\qquad
|\beta_1^\star|=0<|\beta_2^\star|.
\]
This proves that SHAP importance ordering and projection-coefficient ordering need not agree. In particular, a feature may have positive global SHAP importance while having zero projection direction.
\end{proof}

\subsection{Proof of Proposition~\ref{prop:model_response_projection_target}}
\label{app:proof_model_response_projection_target}

\begin{proof}
Consider the population objective
\[
Q(\beta)
=
\mathbb{E}
\left[
\left(
\tilde Z-\beta^\top \tilde X_S
\right)^2
\right].
\]
Expanding the square yields
\[
Q(\beta)
=
\mathbb{E}[\tilde Z^2]
-
2\beta^\top
\mathbb{E}[\tilde X_S\tilde Z]
+
\beta^\top
\mathbb{E}[\tilde X_S\tilde X_S^\top]
\beta.
\]
Let
\[
g_S
=
\mathbb{E}[\tilde X_S\tilde Z],
\qquad
\Sigma_S
=
\mathbb{E}[\tilde X_S\tilde X_S^\top].
\]
Then
\[
Q(\beta)
=
\mathbb{E}[\tilde Z^2]
-
2\beta^\top g_S
+
\beta^\top \Sigma_S\beta.
\]
Since \(\Sigma_S\) is nonsingular and positive definite, \(Q(\beta)\) is strictly convex. Hence its unique minimizer is characterized by the first-order condition
\[
\nabla Q(\beta)
=
-2g_S+2\Sigma_S\beta
=
0.
\]
Solving this equation gives
\[
\beta_S^\star
=
\Sigma_S^{-1}g_S
=
\Sigma_S^{-1}\mathbb{E}[\tilde X_S\tilde Z].
\]

It remains to verify the orthogonality condition. Let
\[
r_S
=
\tilde Z-\beta_S^{\star\top}\tilde X_S.
\]
Then
\[
\begin{aligned}
\mathbb{E}[\tilde X_S r_S]
&=
\mathbb{E}
\left[
\tilde X_S
\left(
\tilde Z-\beta_S^{\star\top}\tilde X_S
\right)
\right] \\
&=
\mathbb{E}[\tilde X_S\tilde Z]
-
\mathbb{E}[\tilde X_S\tilde X_S^\top]\beta_S^\star \\
&=
g_S-\Sigma_S\Sigma_S^{-1}g_S \\
&=0.
\end{aligned}
\]
Thus the projection coefficient is unique and the corresponding residual is orthogonal to the selected covariates.
\end{proof}

\subsection{Proof of Proposition~\ref{prop:fidelity_projection_strength}}
\label{app:proof_fidelity_projection_strength}

\begin{proof}
By Proposition~\ref{prop:model_response_projection_target}, the projection residual
\[
r_S
=
\tilde Z-\beta_S^{\star\top}\tilde X_S
\]
satisfies
\[
\mathbb{E}[\tilde X_S r_S]=0.
\]
Since
\[
p_S^\star(X)
=
\beta_S^{\star\top}\tilde X_S,
\]
we have
\[
\mathbb{E}[p_S^\star(X)r_S]
=
\beta_S^{\star\top}
\mathbb{E}[\tilde X_S r_S]
=
0.
\]
Using the decomposition
\[
\tilde Z=p_S^\star(X)+r_S,
\]
it follows that
\[
\begin{aligned}
\mathbb{E}[\tilde Z^2]
&=
\mathbb{E}
\left[
\left(
p_S^\star(X)+r_S
\right)^2
\right] \\
&=
\mathbb{E}\left[(p_S^\star(X))^2\right]
+
2\mathbb{E}[p_S^\star(X)r_S]
+
\mathbb{E}[r_S^2] \\
&=
\mathbb{E}\left[(p_S^\star(X))^2\right]
+
\mathbb{E}[r_S^2].
\end{aligned}
\]
Therefore,
\[
F_S^\star
=
1-
\frac{\mathbb{E}[r_S^2]}{\mathbb{E}[\tilde Z^2]}
=
\frac{
\mathbb{E}\left[(p_S^\star(X))^2\right]
}{
\mathbb{E}[\tilde Z^2]
}.
\]
Since both terms on the right-hand side of
\[
\mathbb{E}[\tilde Z^2]
=
\mathbb{E}\left[(p_S^\star(X))^2\right]
+
\mathbb{E}[r_S^2]
\]
are nonnegative, we also obtain
\[
0\leq F_S^\star\leq 1.
\]

If \(\operatorname{Var}(p_S^\star(X))>0\), then \(p_S^\star(X)\) is centered because \(\mathbb{E}[\tilde X_S]=0\). Hence
\[
\operatorname{Cov}
\left(
\tilde Z,
p_S^\star(X)
\right)
=
\mathbb{E}[\tilde Zp_S^\star(X)].
\]
Using \(\tilde Z=p_S^\star(X)+r_S\),
\[
\begin{aligned}
\operatorname{Cov}
\left(
\tilde Z,
p_S^\star(X)
\right)
&=
\mathbb{E}
\left[
\left(
p_S^\star(X)+r_S
\right)
p_S^\star(X)
\right] \\
&=
\mathbb{E}\left[(p_S^\star(X))^2\right]
+
\mathbb{E}[p_S^\star(X)r_S] \\
&=
\mathbb{E}\left[(p_S^\star(X))^2\right].
\end{aligned}
\]
Since \(p_S^\star(X)\) is centered,
\[
\operatorname{Var}(p_S^\star(X))
=
\mathbb{E}\left[(p_S^\star(X))^2\right].
\]
Thus
\[
\begin{aligned}
\operatorname{Corr}^2
\left(
\tilde Z,
p_S^\star(X)
\right)
&=
\frac{
\operatorname{Cov}^2(\tilde Z,p_S^\star(X))
}{
\operatorname{Var}(\tilde Z)\operatorname{Var}(p_S^\star(X))
} \\
&=
\frac{
\left\{
\mathbb{E}\left[(p_S^\star(X))^2\right]
\right\}^2
}{
\mathbb{E}[\tilde Z^2]\,
\mathbb{E}\left[(p_S^\star(X))^2\right]
} \\
&=
\frac{
\mathbb{E}\left[(p_S^\star(X))^2\right]
}{
\mathbb{E}[\tilde Z^2]
}
=
F_S^\star.
\end{aligned}
\]

Finally, let
\[
g_S
=
\mathbb{E}[\tilde X_S\tilde Z].
\]
By Proposition~\ref{prop:model_response_projection_target},
\[
\beta_S^\star
=
\Sigma_S^{-1}g_S.
\]
Therefore,
\[
\begin{aligned}
\mathbb{E}
\left[
(p_S^\star(X))^2
\right]
&=
\mathbb{E}
\left[
\left(
\beta_S^{\star\top}\tilde X_S
\right)^2
\right] \\
&=
\beta_S^{\star\top}
\Sigma_S
\beta_S^\star \\
&=
g_S^\top
\Sigma_S^{-1}
\Sigma_S
\Sigma_S^{-1}
g_S \\
&=
g_S^\top
\Sigma_S^{-1}
g_S.
\end{aligned}
\]
When \(\mathbb{E}[\tilde Z^2]=1\), this gives
\[
F_S^\star
=
g_S^\top\Sigma_S^{-1}g_S.
\]
\end{proof}

\subsection{Proof of Proposition~\ref{prop:projection_coefficient_uncertainty}}
\label{app:proof_projection_coefficient_uncertainty}

\begin{proof}
Define
\[
\hat\Sigma_S
=
\frac{1}{m}
\sum_{i=1}^m
\tilde X_{i,S}\tilde X_{i,S}^\top,
\qquad
\hat g_S
=
\frac{1}{m}
\sum_{i=1}^m
\tilde X_{i,S}\tilde Z_i.
\]
The empirical projection coefficient satisfies
\[
\hat\beta_S
=
\hat\Sigma_S^{-1}\hat g_S
\]
whenever \(\hat\Sigma_S\) is nonsingular. By the law of large numbers,
\[
\hat\Sigma_S
\overset{p}{\longrightarrow}
\Sigma_S,
\qquad
\hat g_S
\overset{p}{\longrightarrow}
g_S
=
\mathbb{E}[\tilde X_S\tilde Z].
\]
Since \(\Sigma_S\) is nonsingular, \(\hat\Sigma_S\) is nonsingular with probability approaching one, and the continuous mapping theorem gives
\[
\hat\beta_S
=
\hat\Sigma_S^{-1}\hat g_S
\overset{p}{\longrightarrow}
\Sigma_S^{-1}g_S
=
\beta_S^\star.
\]

Next, write
\[
r_{i,S}
=
\tilde Z_i-\beta_S^{\star\top}\tilde X_{i,S}.
\]
Using
\[
\hat g_S-\hat\Sigma_S\beta_S^\star
=
\frac{1}{m}
\sum_{i=1}^m
\tilde X_{i,S}
\left(
\tilde Z_i-\beta_S^{\star\top}\tilde X_{i,S}
\right),
\]
we obtain
\[
\hat g_S-\hat\Sigma_S\beta_S^\star
=
\frac{1}{m}
\sum_{i=1}^m
\tilde X_{i,S}r_{i,S}.
\]
Therefore,
\[
\begin{aligned}
\hat\beta_S-\beta_S^\star
&=
\hat\Sigma_S^{-1}\hat g_S-\beta_S^\star \\
&=
\hat\Sigma_S^{-1}
\left(
\hat g_S-\hat\Sigma_S\beta_S^\star
\right) \\
&=
\hat\Sigma_S^{-1}
\left(
\frac{1}{m}
\sum_{i=1}^m
\tilde X_{i,S}r_{i,S}
\right).
\end{aligned}
\]
Multiplying by \(\sqrt{m}\) gives
\[
\sqrt{m}
\left(
\hat\beta_S-\beta_S^\star
\right)
=
\hat\Sigma_S^{-1}
\left(
\frac{1}{\sqrt{m}}
\sum_{i=1}^m
\tilde X_{i,S}r_{i,S}
\right).
\]

By Proposition~\ref{prop:model_response_projection_target},
\[
\mathbb{E}[\tilde X_S r_S]=0.
\]
Under the stated fourth-moment assumptions,
\[
\Omega_S
=
\mathbb{E}
\left[
\tilde X_S\tilde X_S^\top r_S^2
\right]
\]
is finite. Hence the multivariate central limit theorem implies
\[
\frac{1}{\sqrt{m}}
\sum_{i=1}^m
\tilde X_{i,S}r_{i,S}
\overset{d}{\longrightarrow}
N(0,\Omega_S).
\]
Since
\[
\hat\Sigma_S^{-1}
\overset{p}{\longrightarrow}
\Sigma_S^{-1},
\]
Slutsky's theorem yields
\[
\sqrt{m}
\left(
\hat\beta_S-\beta_S^\star
\right)
\overset{d}{\longrightarrow}
N
\left(
0,
\Sigma_S^{-1}\Omega_S\Sigma_S^{-1}
\right).
\]
The asymptotic variance of the \(j\)th component is therefore
\[
\frac{
\left[
\Sigma_S^{-1}\Omega_S\Sigma_S^{-1}
\right]_{jj}
}{m},
\]
and the corresponding asymptotic standard error is
\[
\operatorname{se}(\hat\beta_j)
=
\sqrt{
\frac{
\left[
\Sigma_S^{-1}\Omega_S\Sigma_S^{-1}
\right]_{jj}
}{m}
}.
\]
\end{proof}

\subsection{Proof of Proposition~\ref{prop:coefficient_stability_perturbation}}
\label{app:proof_coefficient_stability_perturbation}

\begin{proof}
Since
\[
\left\|
\Sigma_S^{-1}\Delta_{\Sigma}
\right\|_{\mathrm{op}}
<1,
\]
the matrix
\[
I+\Sigma_S^{-1}\Delta_{\Sigma}
\]
is nonsingular. Hence
\[
\Sigma_S+\Delta_{\Sigma}
=
\Sigma_S
\left(
I+\Sigma_S^{-1}\Delta_{\Sigma}
\right)
\]
is also nonsingular.

Using
\[
\beta_S^\star
=
\Sigma_S^{-1}g_S,
\qquad
\beta_S^\Delta
=
(\Sigma_S+\Delta_{\Sigma})^{-1}(g_S+\Delta_g),
\]
we have
\[
g_S=\Sigma_S\beta_S^\star.
\]
Therefore,
\[
\begin{aligned}
\beta_S^\Delta-\beta_S^\star
&=
(\Sigma_S+\Delta_{\Sigma})^{-1}
(g_S+\Delta_g)
-
\beta_S^\star \\
&=
(\Sigma_S+\Delta_{\Sigma})^{-1}
(\Sigma_S\beta_S^\star+\Delta_g)
-
\beta_S^\star \\
&=
(\Sigma_S+\Delta_{\Sigma})^{-1}
\left(
\Sigma_S\beta_S^\star+\Delta_g
-
(\Sigma_S+\Delta_{\Sigma})\beta_S^\star
\right) \\
&=
(\Sigma_S+\Delta_{\Sigma})^{-1}
\left(
\Delta_g-\Delta_{\Sigma}\beta_S^\star
\right).
\end{aligned}
\]

Taking Euclidean norms gives
\[
\left\|
\beta_S^\Delta-\beta_S^\star
\right\|_2
\leq
\left\|
(\Sigma_S+\Delta_{\Sigma})^{-1}
\right\|_{\mathrm{op}}
\left(
\|\Delta_g\|_2
+
\|\Delta_{\Sigma}\|_{\mathrm{op}}
\|\beta_S^\star\|_2
\right).
\]

It remains to bound the inverse term. Since
\[
\Sigma_S+\Delta_{\Sigma}
=
\Sigma_S
\left(
I+\Sigma_S^{-1}\Delta_{\Sigma}
\right),
\]
we can write
\[
(\Sigma_S+\Delta_{\Sigma})^{-1}
=
\left(
I+\Sigma_S^{-1}\Delta_{\Sigma}
\right)^{-1}
\Sigma_S^{-1}.
\]
The Neumann-series bound gives
\[
\left\|
\left(
I+\Sigma_S^{-1}\Delta_{\Sigma}
\right)^{-1}
\right\|_{\mathrm{op}}
\leq
\frac{1}{
1-\|\Sigma_S^{-1}\Delta_{\Sigma}\|_{\mathrm{op}}
}.
\]
Thus
\[
\left\|
(\Sigma_S+\Delta_{\Sigma})^{-1}
\right\|_{\mathrm{op}}
\leq
\frac{
\|\Sigma_S^{-1}\|_{\mathrm{op}}
}{
1-\|\Sigma_S^{-1}\Delta_{\Sigma}\|_{\mathrm{op}}
}.
\]
Substituting this into the previous inequality yields
\[
\left\|
\beta_S^\Delta-\beta_S^\star
\right\|_2
\leq
\frac{
\|\Sigma_S^{-1}\|_{\mathrm{op}}
}{
1-\|\Sigma_S^{-1}\Delta_{\Sigma}\|_{\mathrm{op}}
}
\left(
\|\Delta_g\|_2
+
\|\Delta_{\Sigma}\|_{\mathrm{op}}
\|\beta_S^\star\|_2
\right).
\]
This proves the result.
\end{proof}

\subsection{Proof of Proposition~\ref{prop:fidelity_stability_nonsubstitutable}}
\label{app:proof_fidelity_stability_nonsubstitutable}

\begin{proof}
We prove the two claims by construction.

First, we construct a case with perfect fidelity and unstable coefficient allocation. Let
\[
S=\{1,2\},
\]
and let \((X_1,X_2)\) be centered and standardized with covariance matrix
\[
\Sigma_S
=
\begin{pmatrix}
1 & \rho \\
\rho & 1
\end{pmatrix},
\qquad
0<\rho<1.
\]
Define the fitted response
\[
Z=f(X)=X_1+X_2.
\]
Since \(Z\) lies in the linear span of \(X_S\), its projection onto the selected features has zero residual. Hence
\[
F_S^\star=1.
\]

Now perturb the feature-response moment \(g_S=\mathbb{E}[\tilde X_S\tilde Z]\) by
\[
\Delta_g
=
\delta
\begin{pmatrix}
1\\
-1
\end{pmatrix},
\qquad
\delta>0,
\]
while keeping \(\Sigma_S\) fixed. The resulting coefficient perturbation is
\[
\beta_S^\Delta-\beta_S^\star
=
\Sigma_S^{-1}\Delta_g.
\]
Since
\[
\Sigma_S^{-1}
=
\frac{1}{1-\rho^2}
\begin{pmatrix}
1 & -\rho \\
-\rho & 1
\end{pmatrix},
\]
we have
\[
\Sigma_S^{-1}
\begin{pmatrix}
1\\
-1
\end{pmatrix}
=
\frac{1}{1-\rho}
\begin{pmatrix}
1\\
-1
\end{pmatrix}.
\]
Therefore,
\[
\left\|
\beta_S^\Delta-\beta_S^\star
\right\|_2
=
\frac{
\|\Delta_g\|_2
}{
1-\rho
}.
\]
For any \(M>0\), choose \(\rho\) sufficiently close to one so that
\[
\frac{1}{1-\rho}>M.
\]
Then the coefficient perturbation is amplified by more than \(M\), even though \(F_S^\star=1\). This proves that perfect fidelity can coexist with arbitrarily poor coefficient stability.

Second, we construct a case with well-conditioned coefficients and arbitrarily low fidelity. Let
\[
S = \{1\},
\]
and let \(X\) be a centered, standardized, symmetric random variable with finite fourth moment. Define
\[
U = X^2 - E(X^2),
\]
and consider
\[
Z = f(X) = aX + bU, \qquad a \neq 0.
\]
By symmetry,
\[
E[XU] = E\left[X\{X^2 - E(X^2)\}\right] = 0.
\]
Thus the linear and quadratic components are orthogonal. Since \(\Sigma_S = 1\), the selected-feature covariance is perfectly conditioned and perturbations of \(g_S\) are not amplified by the feature covariance.

The population fidelity is the fraction of fitted-response variance captured by the linear projection. Because \(X\) and \(U\) are orthogonal and \(\operatorname{Var}(X)=1\),
\[
\operatorname{Var}(Z)=a^2+b^2\operatorname{Var}(U).
\]
The projected component is proportional to \(aX\), so
\[
F_S^\star =
\frac{
a^2
}{
a^2+b^2\operatorname{Var}(U)
}.
\]
For any \(\eta>0\), choose \(|b|\) sufficiently large so that
\[
\frac{
a^2
}{
a^2+b^2\operatorname{Var}(U)
}
<\eta.
\]
Hence the coefficient map is well-conditioned while the projection fidelity is arbitrarily small.

The two constructions show that high fidelity does not imply coefficient stability, and coefficient stability does not imply high fidelity.
\end{proof}

\section{Robustness and Sensitivity Analyses}
\label{app:robustness_sensitivity}

\subsection{Common Design}
\label{app:common_sensitivity_design}

These appendix analyses examine whether the proposed table depends strongly on specific experimental choices. We vary the number of selected features, the explanation-sample size, the SHAP-ranking sample size, the black-box backbone, and the Shapley-based global ranker used to select the feature set. Across all analyses, the explanation target is the fitted black-box response $f(X)$ rather than the observed outcome $Y$. After a feature set is selected, the $\phi$-table is always fit in the same way: the selected covariates are standardized, the fitted response is standardized, and the surrogate regression is estimated with HC3 robust standard errors. Thus, the sensitivity experiments change the upstream source of the selected feature set or the available explanation sample, while keeping the statistical table construction fixed.

For the top-$k$, explanation-sample-size, and SHAP-sample-size analyses, we use the seven tabular regression datasets from the main real-data experiments: Air Quality, Airfoil Self-Noise, Bike Sharing, California Housing, Combined Cycle Power Plant, Concrete Compressive Strength, and Energy Efficiency. Unless otherwise stated, the black-box model is a random forest regressor. We compute a global SHAP ranking on the explanation pool, select the SHAP-ranked features under the corresponding experimental condition, and then fit the $\phi$-table to $f(X)$. The preprocessing pipeline follows the main experiments: usable numeric covariates are retained, leakage columns are removed where applicable, missing values are median-imputed, constant features are discarded, covariates are standardized, and the target is standardized before black-box training. Results are averaged over 20 random splits, with coefficient stability computed from 40 bootstrap resamples within each split.

The backbone robustness analysis changes the fitted black-box model while keeping the SHAP-based selection and table construction otherwise fixed. We replace the random forest with histogram gradient boosting, multilayer perceptron regression, or RBF-kernel support vector regression. Because model-agnostic permutation SHAP is substantially more expensive for non-tree backbones, this analysis is run on Bike Sharing, California Housing, Concrete Compressive Strength, and Energy Efficiency with 10 random splits and 30 bootstrap resamples per split. We additionally report ModelR2 in this analysis to separate black-box predictive performance from surrogate fidelity: ModelR2 measures how well the black-box predicts the original target, whereas Fid measures how well the selected-feature surrogate summarizes the fitted response $f(X)$.

The final robustness analysis changes the Shapley-based global ranker rather than the backbone. To avoid tying the comparison to tree-specific SHAP implementations, we use an MLP backbone and compute feature rankings using six Shapley-based variants: interventional Monte Carlo SHAP, conditional SHAP with a Gaussian conditional approximation, SAGE, Shapley effects, Baseline Shapley, and a second-order Shapley--Taylor feature reduction. Each ranker selects its top five features, after which the same $\phi$-table procedure is applied. This analysis is run on Air Quality, Bike Sharing, California Housing, Concrete Compressive Strength, and Energy Efficiency; Airfoil Self-Noise and Combined Cycle Power Plant are excluded because, under $k=5$, the selected set nearly exhausts the usable feature set and ranker differences become mechanically limited. In addition to the standard table diagnostics, we report top-$k$ overlap with the SHAP-selected set to distinguish agreement in feature selection from agreement in the resulting statistical interpretation.

\subsection{Sensitivity to the Number of Selected Features}
\label{app:sensitivity_topk}

\paragraph{Experimental setup.}
This experiment varies the number of SHAP-selected features used to construct the surrogate table. We set $k \in \{3,5,7\}$ and keep the rest of the random-forest explanation pipeline fixed. When a dataset has fewer usable features than the requested value of $k$, the selected set is capped at the available feature dimension.

\paragraph{Results and analysis.}
Table~\ref{tab:app_topk_sensitivity} shows that increasing $k$ usually improves surrogate fidelity, with the largest change often occurring between $k=3$ and $k=5$. AFN, CAH, CCS, and EE follow this pattern, indicating that the top-three table can omit useful variation in the fitted response. The additional features do not always sharpen the coefficient-level reading: in Bike and CCS, larger $k$ improves descriptive fit while increasing ambiguity or coefficient dispersion. CAH behaves differently, as larger selected sets reduce confidence interval width and coefficient dispersion, suggesting that very small tables can under-specify the model-response projection. The sweep therefore treats $k$ as an interpretive size choice: larger tables can capture more of $f(X)$, but fidelity gains should be read together with ambiguity, uncertainty, and bootstrap stability.

\begin{table*}[ht]
\centering
\caption{Top-$k$ sensitivity of the statistical explanation table. Values are mean $\pm$ standard deviation over repeated splits.}
\label{tab:app_topk_sensitivity}
\resizebox{\textwidth}{!}{
\begin{tabular}{llccccccc}
\toprule
Dataset & $k$ & Actual $k$ & Fid & DirRate & Ambig@$k$ & CIW & BStab & BDisp \\
\midrule
\multirow{3}{*}{AQ}
& 3 & 3 & $0.955 \pm 0.004$ & $0.933 \pm 0.137$ & $0.067 \pm 0.137$ & $0.117 \pm 0.011$ & $0.942 \pm 0.009$ & $0.062 \pm 0.010$ \\
& 5 & 5 & $0.958 \pm 0.005$ & $0.710 \pm 0.152$ & $0.290 \pm 0.152$ & $0.090 \pm 0.009$ & $0.938 \pm 0.007$ & $0.066 \pm 0.008$ \\
& 7 & 7 & $0.961 \pm 0.004$ & $0.814 \pm 0.094$ & $0.186 \pm 0.094$ & $0.081 \pm 0.008$ & $0.933 \pm 0.010$ & $0.072 \pm 0.012$ \\
\midrule
\multirow{3}{*}{AFN}
& 3 & 3 & $0.556 \pm 0.022$ & $1.000 \pm 0.000$ & $0.000 \pm 0.000$ & $0.137 \pm 0.010$ & $0.945 \pm 0.006$ & $0.059 \pm 0.006$ \\
& 5 & 5 & $0.630 \pm 0.019$ & $1.000 \pm 0.000$ & $0.000 \pm 0.000$ & $0.153 \pm 0.007$ & $0.928 \pm 0.005$ & $0.078 \pm 0.006$ \\
& 7 & 5 & $0.630 \pm 0.019$ & $1.000 \pm 0.000$ & $0.000 \pm 0.000$ & $0.153 \pm 0.007$ & $0.926 \pm 0.005$ & $0.080 \pm 0.006$ \\
\midrule
\multirow{3}{*}{Bike}
& 3 & 3 & $0.427 \pm 0.015$ & $1.000 \pm 0.000$ & $0.000 \pm 0.000$ & $0.060 \pm 0.001$ & $0.962 \pm 0.002$ & $0.040 \pm 0.003$ \\
& 5 & 5 & $0.430 \pm 0.015$ & $0.740 \pm 0.131$ & $0.260 \pm 0.131$ & $0.139 \pm 0.081$ & $0.883 \pm 0.067$ & $0.139 \pm 0.088$ \\
& 7 & 7 & $0.456 \pm 0.015$ & $0.700 \pm 0.079$ & $0.300 \pm 0.079$ & $0.156 \pm 0.030$ & $0.824 \pm 0.032$ & $0.216 \pm 0.050$ \\
\midrule
\multirow{3}{*}{CAH}
& 3 & 3 & $0.738 \pm 0.023$ & $0.783 \pm 0.196$ & $0.217 \pm 0.196$ & $2.598 \pm 1.764$ & $0.676 \pm 0.228$ & $0.728 \pm 0.810$ \\
& 5 & 5 & $0.808 \pm 0.019$ & $0.880 \pm 0.101$ & $0.120 \pm 0.101$ & $1.572 \pm 1.025$ & $0.698 \pm 0.191$ & $0.566 \pm 0.523$ \\
& 7 & 7 & $0.828 \pm 0.016$ & $0.914 \pm 0.072$ & $0.086 \pm 0.072$ & $1.125 \pm 0.693$ & $0.729 \pm 0.168$ & $0.455 \pm 0.382$ \\
\midrule
\multirow{3}{*}{CCPP}
& 3 & 3 & $0.976 \pm 0.001$ & $1.000 \pm 0.000$ & $0.000 \pm 0.000$ & $0.025 \pm 0.001$ & $0.989 \pm 0.001$ & $0.011 \pm 0.001$ \\
& 5 & 4 & $0.976 \pm 0.001$ & $1.000 \pm 0.000$ & $0.000 \pm 0.000$ & $0.024 \pm 0.001$ & $0.988 \pm 0.001$ & $0.012 \pm 0.001$ \\
& 7 & 4 & $0.976 \pm 0.001$ & $1.000 \pm 0.000$ & $0.000 \pm 0.000$ & $0.024 \pm 0.001$ & $0.988 \pm 0.001$ & $0.012 \pm 0.001$ \\
\midrule
\multirow{3}{*}{CCS}
& 3 & 3 & $0.491 \pm 0.032$ & $1.000 \pm 0.000$ & $0.000 \pm 0.000$ & $0.177 \pm 0.012$ & $0.910 \pm 0.007$ & $0.099 \pm 0.008$ \\
& 5 & 5 & $0.613 \pm 0.025$ & $1.000 \pm 0.000$ & $0.000 \pm 0.000$ & $0.173 \pm 0.009$ & $0.898 \pm 0.007$ & $0.114 \pm 0.009$ \\
& 7 & 7 & $0.627 \pm 0.024$ & $0.900 \pm 0.082$ & $0.100 \pm 0.082$ & $0.208 \pm 0.012$ & $0.864 \pm 0.014$ & $0.158 \pm 0.018$ \\
\midrule
\multirow{3}{*}{EE}
& 3 & 3 & $0.881 \pm 0.027$ & $0.867 \pm 0.199$ & $0.133 \pm 0.199$ & $0.266 \pm 0.150$ & $0.917 \pm 0.047$ & $0.093 \pm 0.055$ \\
& 5 & 5 & $0.925 \pm 0.005$ & $0.980 \pm 0.062$ & $0.020 \pm 0.062$ & $0.275 \pm 0.040$ & $0.888 \pm 0.014$ & $0.126 \pm 0.017$ \\
& 7 & 7 & $0.927 \pm 0.005$ & $0.886 \pm 0.075$ & $0.114 \pm 0.075$ & $0.205 \pm 0.013$ & $0.900 \pm 0.008$ & $0.111 \pm 0.009$ \\
\bottomrule
\end{tabular}
}
\end{table*}

\subsection{Sensitivity to the Explanation Sample Size}
\label{app:sensitivity_explanation_sample_size}

\paragraph{Experimental setup.}
This experiment fixes the selected feature set to the SHAP top-$5$ features and varies the number of explanation observations used to estimate the surrogate coefficients. We consider $n \in \{100,200,300\}$ while keeping the random-forest model, SHAP ranking procedure, and table diagnostics unchanged.

\paragraph{Results and analysis.}
Table~\ref{tab:app_explanation_sample_size_sensitivity} shows that larger explanation samples mainly improve estimation precision. Confidence intervals become narrower, coefficient dispersion decreases, and bootstrap stability generally increases as $n$ grows. Directional ambiguity also falls in AFN, CCS, and EE, showing that some weak directional signals at small $n$ are estimation-driven. Fidelity remains comparatively stable because it reflects how well the fixed top-$5$ feature set can linearly summarize $f(X)$, not merely how many rows are used to estimate the projection. Bike remains difficult even at larger $n$, which separates sampling precision from the structural adequacy of the selected linear surrogate.

\begin{table*}[ht]
\centering
\caption{Explanation-sample-size sensitivity of the statistical explanation table. Values are mean $\pm$ standard deviation over repeated splits.}
\label{tab:app_explanation_sample_size_sensitivity}
\resizebox{\textwidth}{!}{
\begin{tabular}{lccccccc}
\toprule
Dataset & $n$ & Fid & DirRate & Ambig@$k$ & CIW & BStab & BDisp \\
\midrule
\multirow{3}{*}{AQ}
& 100 & $0.968 \pm 0.017$ & $0.620 \pm 0.233$ & $0.380 \pm 0.233$ & $0.379 \pm 0.153$ & $0.817 \pm 0.059$ & $0.231 \pm 0.090$ \\
& 200 & $0.961 \pm 0.015$ & $0.670 \pm 0.163$ & $0.330 \pm 0.163$ & $0.302 \pm 0.084$ & $0.832 \pm 0.041$ & $0.205 \pm 0.058$ \\
& 300 & $0.959 \pm 0.011$ & $0.670 \pm 0.175$ & $0.330 \pm 0.175$ & $0.276 \pm 0.065$ & $0.845 \pm 0.041$ & $0.187 \pm 0.060$ \\
\midrule
\multirow{3}{*}{AFN}
& 100 & $0.639 \pm 0.054$ & $0.840 \pm 0.154$ & $0.160 \pm 0.154$ & $0.418 \pm 0.060$ & $0.829 \pm 0.027$ & $0.208 \pm 0.040$ \\
& 200 & $0.637 \pm 0.048$ & $0.980 \pm 0.062$ & $0.020 \pm 0.062$ & $0.269 \pm 0.021$ & $0.877 \pm 0.015$ & $0.140 \pm 0.020$ \\
& 300 & $0.639 \pm 0.035$ & $0.990 \pm 0.045$ & $0.010 \pm 0.045$ & $0.216 \pm 0.015$ & $0.898 \pm 0.010$ & $0.114 \pm 0.012$ \\
\midrule
\multirow{3}{*}{Bike}
& 100 & $0.448 \pm 0.069$ & $0.520 \pm 0.136$ & $0.480 \pm 0.136$ & $0.840 \pm 0.529$ & $0.669 \pm 0.147$ & $0.573 \pm 0.387$ \\
& 200 & $0.434 \pm 0.043$ & $0.560 \pm 0.139$ & $0.440 \pm 0.139$ & $0.507 \pm 0.344$ & $0.731 \pm 0.127$ & $0.417 \pm 0.300$ \\
& 300 & $0.433 \pm 0.033$ & $0.570 \pm 0.098$ & $0.430 \pm 0.098$ & $0.368 \pm 0.227$ & $0.776 \pm 0.096$ & $0.309 \pm 0.172$ \\
\midrule
\multirow{3}{*}{CAH}
& 100 & $0.843 \pm 0.040$ & $0.870 \pm 0.149$ & $0.130 \pm 0.149$ & $1.634 \pm 0.852$ & $0.771 \pm 0.147$ & $0.362 \pm 0.358$ \\
& 200 & $0.836 \pm 0.028$ & $0.930 \pm 0.098$ & $0.070 \pm 0.098$ & $1.372 \pm 1.125$ & $0.820 \pm 0.170$ & $0.311 \pm 0.471$ \\
& 300 & $0.829 \pm 0.027$ & $0.930 \pm 0.098$ & $0.070 \pm 0.098$ & $1.340 \pm 1.170$ & $0.812 \pm 0.163$ & $0.308 \pm 0.410$ \\
\midrule
\multirow{3}{*}{CCPP}
& 100 & $0.977 \pm 0.005$ & $0.775 \pm 0.112$ & $0.225 \pm 0.112$ & $0.127 \pm 0.028$ & $0.940 \pm 0.012$ & $0.063 \pm 0.014$ \\
& 200 & $0.977 \pm 0.003$ & $0.800 \pm 0.103$ & $0.200 \pm 0.103$ & $0.088 \pm 0.018$ & $0.958 \pm 0.008$ & $0.044 \pm 0.009$ \\
& 300 & $0.977 \pm 0.002$ & $0.812 \pm 0.111$ & $0.188 \pm 0.111$ & $0.068 \pm 0.008$ & $0.967 \pm 0.004$ & $0.034 \pm 0.004$ \\
\midrule
\multirow{3}{*}{CCS}
& 100 & $0.649 \pm 0.067$ & $0.830 \pm 0.163$ & $0.170 \pm 0.163$ & $0.399 \pm 0.077$ & $0.804 \pm 0.036$ & $0.247 \pm 0.057$ \\
& 200 & $0.628 \pm 0.054$ & $0.960 \pm 0.082$ & $0.040 \pm 0.082$ & $0.262 \pm 0.024$ & $0.859 \pm 0.012$ & $0.164 \pm 0.016$ \\
& 300 & $0.609 \pm 0.029$ & $1.000 \pm 0.000$ & $0.000 \pm 0.000$ & $0.209 \pm 0.017$ & $0.877 \pm 0.008$ & $0.140 \pm 0.011$ \\
\midrule
\multirow{3}{*}{EE}
& 100 & $0.937 \pm 0.013$ & $0.910 \pm 0.102$ & $0.090 \pm 0.102$ & $0.518 \pm 0.051$ & $0.813 \pm 0.028$ & $0.231 \pm 0.043$ \\
& 200 & $0.934 \pm 0.007$ & $0.980 \pm 0.062$ & $0.020 \pm 0.062$ & $0.343 \pm 0.030$ & $0.876 \pm 0.015$ & $0.142 \pm 0.020$ \\
& 300 & $0.932 \pm 0.007$ & $0.990 \pm 0.045$ & $0.010 \pm 0.045$ & $0.275 \pm 0.018$ & $0.893 \pm 0.014$ & $0.120 \pm 0.017$ \\
\bottomrule
\end{tabular}
}
\end{table*}

\subsection{Sensitivity to the SHAP Sample Size}
\label{app:sensitivity_shap_sample_size}

\paragraph{Experimental setup.}
This experiment varies the number of observations used to estimate the global SHAP ranking. We compute SHAP rankings from samples of size $m \in \{50,100,300\}$ and compare each ranking with a reference ranking computed from up to 600 explanation-pool observations. The table itself is then constructed with a fixed explanation sample size of 300 observations. In addition to the standard diagnostics, we report top-$k$ overlap with the reference selected set (TopKOv) and Spearman rank correlation with the reference SHAP importance vector (RankCorr).

\paragraph{Results and analysis.}
Table~\ref{tab:app_shap_sample_size_sensitivity} shows that the SHAP ranking is already stable at small sample sizes in this setting. TopKOv and RankCorr are high even at $m=50$, and increasing $m$ produces only modest changes for most datasets. The downstream diagnostics also remain nearly unchanged because the selected feature sets change little. Bike illustrates the main distinction: its SHAP ranking is stable, but its fidelity and ambiguity remain weaker than in several other datasets. Ranking stability therefore does not substitute for coefficient-level diagnostics; it only confirms that the selected features are not an artifact of a fragile SHAP sampling choice.

\begin{table*}[ht]
\centering
\caption{SHAP-sample-size sensitivity of the statistical explanation table. Values are mean $\pm$ standard deviation over repeated splits.}
\label{tab:app_shap_sample_size_sensitivity}
\resizebox{\textwidth}{!}{
\begin{tabular}{llcccccccc}
\toprule
Dataset & $m$ & TopKOv & RankCorr & Fid & DirRate & Ambig@$k$ & CIW & BStab & BDisp \\
\midrule
\multirow{3}{*}{AQ}
& 50  & $0.920 \pm 0.101$ & $0.960 \pm 0.031$ & $0.960 \pm 0.015$ & $0.710 \pm 0.165$ & $0.290 \pm 0.165$ & $0.248 \pm 0.068$ & $0.849 \pm 0.052$ & $0.183 \pm 0.075$ \\
& 100 & $0.950 \pm 0.089$ & $0.977 \pm 0.019$ & $0.960 \pm 0.015$ & $0.710 \pm 0.165$ & $0.290 \pm 0.165$ & $0.246 \pm 0.069$ & $0.851 \pm 0.052$ & $0.179 \pm 0.073$ \\
& 300 & $0.940 \pm 0.094$ & $0.981 \pm 0.025$ & $0.960 \pm 0.015$ & $0.720 \pm 0.151$ & $0.280 \pm 0.151$ & $0.249 \pm 0.066$ & $0.852 \pm 0.047$ & $0.177 \pm 0.066$ \\
\midrule
\multirow{3}{*}{AFN}
& 50  & $1.000 \pm 0.000$ & $1.000 \pm 0.000$ & $0.637 \pm 0.039$ & $1.000 \pm 0.000$ & $0.000 \pm 0.000$ & $0.221 \pm 0.018$ & $0.897 \pm 0.010$ & $0.115 \pm 0.013$ \\
& 100 & $1.000 \pm 0.000$ & $1.000 \pm 0.000$ & $0.637 \pm 0.039$ & $1.000 \pm 0.000$ & $0.000 \pm 0.000$ & $0.221 \pm 0.018$ & $0.896 \pm 0.011$ & $0.116 \pm 0.014$ \\
& 300 & $1.000 \pm 0.000$ & $1.000 \pm 0.000$ & $0.637 \pm 0.039$ & $1.000 \pm 0.000$ & $0.000 \pm 0.000$ & $0.221 \pm 0.018$ & $0.900 \pm 0.012$ & $0.112 \pm 0.014$ \\
\midrule
\multirow{3}{*}{Bike}
& 50  & $0.970 \pm 0.073$ & $0.990 \pm 0.011$ & $0.450 \pm 0.035$ & $0.570 \pm 0.163$ & $0.430 \pm 0.163$ & $0.447 \pm 0.301$ & $0.768 \pm 0.105$ & $0.326 \pm 0.190$ \\
& 100 & $1.000 \pm 0.000$ & $0.991 \pm 0.012$ & $0.448 \pm 0.034$ & $0.550 \pm 0.157$ & $0.450 \pm 0.157$ & $0.467 \pm 0.295$ & $0.760 \pm 0.102$ & $0.338 \pm 0.183$ \\
& 300 & $1.000 \pm 0.000$ & $0.998 \pm 0.003$ & $0.448 \pm 0.034$ & $0.550 \pm 0.157$ & $0.450 \pm 0.157$ & $0.467 \pm 0.295$ & $0.759 \pm 0.106$ & $0.343 \pm 0.193$ \\
\midrule
\multirow{3}{*}{CAH}
& 50  & $0.980 \pm 0.062$ & $0.990 \pm 0.012$ & $0.838 \pm 0.016$ & $0.970 \pm 0.073$ & $0.030 \pm 0.073$ & $0.829 \pm 0.232$ & $0.896 \pm 0.031$ & $0.118 \pm 0.041$ \\
& 100 & $1.000 \pm 0.000$ & $0.998 \pm 0.007$ & $0.839 \pm 0.013$ & $0.990 \pm 0.045$ & $0.010 \pm 0.045$ & $0.815 \pm 0.212$ & $0.895 \pm 0.036$ & $0.119 \pm 0.047$ \\
& 300 & $1.000 \pm 0.000$ & $0.999 \pm 0.005$ & $0.839 \pm 0.013$ & $0.990 \pm 0.045$ & $0.010 \pm 0.045$ & $0.815 \pm 0.212$ & $0.893 \pm 0.032$ & $0.121 \pm 0.043$ \\
\midrule
\multirow{3}{*}{CCPP}
& 50  & $1.000 \pm 0.000$ & $1.000 \pm 0.000$ & $0.976 \pm 0.002$ & $0.838 \pm 0.122$ & $0.163 \pm 0.122$ & $0.067 \pm 0.004$ & $0.966 \pm 0.004$ & $0.035 \pm 0.004$ \\
& 100 & $1.000 \pm 0.000$ & $0.990 \pm 0.045$ & $0.976 \pm 0.002$ & $0.838 \pm 0.122$ & $0.163 \pm 0.122$ & $0.067 \pm 0.004$ & $0.968 \pm 0.004$ & $0.033 \pm 0.004$ \\
& 300 & $1.000 \pm 0.000$ & $0.990 \pm 0.045$ & $0.976 \pm 0.002$ & $0.838 \pm 0.122$ & $0.163 \pm 0.122$ & $0.067 \pm 0.004$ & $0.967 \pm 0.003$ & $0.034 \pm 0.003$ \\
\midrule
\multirow{3}{*}{CCS}
& 50  & $1.000 \pm 0.000$ & $0.992 \pm 0.014$ & $0.623 \pm 0.030$ & $0.990 \pm 0.045$ & $0.010 \pm 0.045$ & $0.211 \pm 0.022$ & $0.880 \pm 0.012$ & $0.137 \pm 0.016$ \\
& 100 & $1.000 \pm 0.000$ & $0.990 \pm 0.014$ & $0.623 \pm 0.030$ & $0.990 \pm 0.045$ & $0.010 \pm 0.045$ & $0.211 \pm 0.022$ & $0.881 \pm 0.010$ & $0.135 \pm 0.013$ \\
& 300 & $1.000 \pm 0.000$ & $0.998 \pm 0.007$ & $0.623 \pm 0.030$ & $0.990 \pm 0.045$ & $0.010 \pm 0.045$ & $0.211 \pm 0.022$ & $0.881 \pm 0.009$ & $0.135 \pm 0.011$ \\
\midrule
\multirow{3}{*}{EE}
& 50  & $0.990 \pm 0.045$ & $0.992 \pm 0.018$ & $0.933 \pm 0.007$ & $0.970 \pm 0.073$ & $0.030 \pm 0.073$ & $0.266 \pm 0.027$ & $0.900 \pm 0.011$ & $0.112 \pm 0.014$ \\
& 100 & $1.000 \pm 0.000$ & $0.989 \pm 0.018$ & $0.933 \pm 0.007$ & $0.980 \pm 0.062$ & $0.020 \pm 0.062$ & $0.273 \pm 0.018$ & $0.895 \pm 0.011$ & $0.117 \pm 0.013$ \\
& 300 & $1.000 \pm 0.000$ & $0.998 \pm 0.007$ & $0.933 \pm 0.007$ & $0.980 \pm 0.062$ & $0.020 \pm 0.062$ & $0.273 \pm 0.018$ & $0.896 \pm 0.012$ & $0.116 \pm 0.015$ \\
\bottomrule
\end{tabular}
}
\end{table*}

\subsection{Robustness to Non-Random-Forest Backbones}
\label{app:sensitivity_non_rf_backbone}

\paragraph{Experimental setup.}
This experiment replaces the random-forest black-box model with HGB, MLP, and SVR-RBF backbones. For each fitted model, we compute a model-agnostic permutation SHAP ranking, select the top-$5$ features, and fit the same surrogate table to the standardized model response. ModelR2 is included to separate black-box predictive performance from surrogate fidelity.

\paragraph{Results and analysis.}
Table~\ref{tab:app_non_rf_backbone_robustness} shows that the table construction extends beyond random forests. The diagnostics also vary across backbones, as expected for a method that explains the fitted response $f(X)$. ModelR2 and Fid capture different objects: in Bike, HGB predicts the outcome well but is poorly summarized by the top-$5$ linear surrogate, whereas SVR-RBF has lower predictive performance but higher surrogate fidelity. CAH, CCS, and EE give stronger table-level summaries, with several backbones showing moderate to high fidelity, low ambiguity, and stable bootstrap coefficients. The backbone experiment therefore supports a precise robustness claim: the procedure is applicable across model classes, while the resulting coefficient-level explanation remains specific to the fitted black-box model.

\begin{table*}[ht]
\centering
\caption{Robustness of the statistical explanation table to non-random-forest black-box backbones. Values are mean $\pm$ standard deviation over repeated splits.}
\label{tab:app_non_rf_backbone_robustness}
\resizebox{\textwidth}{!}{
\begin{tabular}{llccccccc}
\toprule
Dataset & Backbone & ModelR2 & Fid & DirRate & Ambig@$k$ & CIW & BStab & BDisp \\
\midrule
\multirow{3}{*}{Bike}
& HGB     & $0.931 \pm 0.005$ & $0.436 \pm 0.031$ & $0.740 \pm 0.165$ & $0.260 \pm 0.165$ & $0.226 \pm 0.176$ & $0.846 \pm 0.078$ & $0.195 \pm 0.143$ \\
& MLP     & $0.781 \pm 0.042$ & $0.443 \pm 0.073$ & $0.580 \pm 0.175$ & $0.420 \pm 0.175$ & $0.503 \pm 0.288$ & $0.726 \pm 0.134$ & $0.422 \pm 0.265$ \\
& SVR-RBF & $0.596 \pm 0.018$ & $0.590 \pm 0.050$ & $0.680 \pm 0.193$ & $0.320 \pm 0.193$ & $0.568 \pm 0.315$ & $0.704 \pm 0.121$ & $0.458 \pm 0.241$ \\
\midrule
\multirow{3}{*}{CAH}
& HGB     & $0.799 \pm 0.011$ & $0.768 \pm 0.029$ & $0.800 \pm 0.094$ & $0.200 \pm 0.094$ & $0.310 \pm 0.064$ & $0.901 \pm 0.009$ & $0.109 \pm 0.012$ \\
& MLP     & $0.742 \pm 0.012$ & $0.842 \pm 0.040$ & $0.880 \pm 0.140$ & $0.120 \pm 0.140$ & $0.244 \pm 0.091$ & $0.926 \pm 0.016$ & $0.080 \pm 0.019$ \\
& SVR-RBF & $0.756 \pm 0.012$ & $0.798 \pm 0.041$ & $0.780 \pm 0.148$ & $0.220 \pm 0.148$ & $0.342 \pm 0.086$ & $0.899 \pm 0.029$ & $0.114 \pm 0.036$ \\
\midrule
\multirow{3}{*}{CCS}
& HGB     & $0.911 \pm 0.009$ & $0.617 \pm 0.028$ & $1.000 \pm 0.000$ & $0.000 \pm 0.000$ & $0.203 \pm 0.019$ & $0.886 \pm 0.013$ & $0.129 \pm 0.017$ \\
& MLP     & $0.873 \pm 0.027$ & $0.674 \pm 0.024$ & $1.000 \pm 0.000$ & $0.000 \pm 0.000$ & $0.189 \pm 0.017$ & $0.912 \pm 0.013$ & $0.096 \pm 0.016$ \\
& SVR-RBF & $0.864 \pm 0.018$ & $0.697 \pm 0.032$ & $1.000 \pm 0.000$ & $0.000 \pm 0.000$ & $0.194 \pm 0.014$ & $0.913 \pm 0.010$ & $0.096 \pm 0.012$ \\
\midrule
\multirow{3}{*}{EE}
& HGB     & $0.996 \pm 0.001$ & $0.903 \pm 0.008$ & $0.820 \pm 0.063$ & $0.180 \pm 0.063$ & $0.289 \pm 0.017$ & $0.941 \pm 0.006$ & $0.063 \pm 0.007$ \\
& MLP     & $0.991 \pm 0.003$ & $0.929 \pm 0.007$ & $0.940 \pm 0.097$ & $0.060 \pm 0.097$ & $0.299 \pm 0.082$ & $0.885 \pm 0.031$ & $0.131 \pm 0.040$ \\
& SVR-RBF & $0.983 \pm 0.005$ & $0.947 \pm 0.006$ & $1.000 \pm 0.000$ & $0.000 \pm 0.000$ & $0.306 \pm 0.039$ & $0.887 \pm 0.013$ & $0.127 \pm 0.017$ \\
\bottomrule
\end{tabular}
}
\end{table*}

\subsection{Robustness across Shapley Global Rankers}
\label{app:shapley_ranker_robustness}

\paragraph{Experimental setup.}
We test whether the $\phi$-table remains useful when the selected feature set is supplied by different Shapley global rankers. Using the MLP setting described in Appendix~\ref{app:common_sensitivity_design}, we compare six rankers: interventional Monte Carlo SHAP, conditional SHAP, SAGE, Shapley effects, Baseline Shapley, and a second-order Shapley--Taylor feature reduction. Each ranker selects its top five features, and the same $\phi$-table procedure is then applied to the fitted response $f(X)$. We report overlap with the SHAP-selected set, surrogate fidelity, directional evidence, ambiguity, interval width, and bootstrap coefficient stability.

\paragraph{Results and analysis.}
Table~\ref{tab:shapley_ranker_robustness} shows that different Shapley rankers can select noticeably different feature sets, but the $\phi$-table remains applicable without ranker-specific changes. The overlap column confirms that several alternatives do not simply reproduce the SHAP top-$k$ set. The remaining diagnostics show why this matters: similar rankings can differ in ambiguity, interval width, and bootstrap dispersion, while lower-overlap rankings can still yield stable or high-fidelity summaries. Thus, the experiment supports the use of the $\phi$-table as a common diagnostic layer for evaluating the directional and statistical strength of a selected Shapley explanation.

\begin{table}[ht]
\centering
\scriptsize
\caption{Robustness of the $\phi$-table across Shapley global rankers with an MLP backbone. All results are reported as mean $\pm$ standard deviation over repeated splits with $k=5$.}
\label{tab:shapley_ranker_robustness}
\resizebox{\textwidth}{!}{
\begin{tabular}{llccccccc}
\toprule
Dataset & Ranker & Ov@k & Fid & DirRate & Ambig@k & CIW & BStab & BDisp \\
\midrule
\multirow{6}{*}{AQ}
& SHAP     & $1.000 \pm 0.000$ & $0.937 \pm 0.009$ & $0.760 \pm 0.184$ & $0.240 \pm 0.184$ & $0.220 \pm 0.038$ & $0.853 \pm 0.028$ & $0.173 \pm 0.039$ \\
& CondSHAP & $0.520 \pm 0.193$ & $0.914 \pm 0.038$ & $0.860 \pm 0.165$ & $0.140 \pm 0.165$ & $0.148 \pm 0.038$ & $0.913 \pm 0.030$ & $0.097 \pm 0.037$ \\
& SAGE     & $0.760 \pm 0.084$ & $0.937 \pm 0.011$ & $0.720 \pm 0.140$ & $0.280 \pm 0.140$ & $0.236 \pm 0.032$ & $0.835 \pm 0.029$ & $0.198 \pm 0.041$ \\
& ShapEff  & $0.760 \pm 0.084$ & $0.937 \pm 0.010$ & $0.800 \pm 0.133$ & $0.200 \pm 0.133$ & $0.233 \pm 0.031$ & $0.847 \pm 0.028$ & $0.182 \pm 0.039$ \\
& BShap    & $0.800 \pm 0.094$ & $0.940 \pm 0.009$ & $0.820 \pm 0.175$ & $0.180 \pm 0.175$ & $0.235 \pm 0.030$ & $0.851 \pm 0.024$ & $0.176 \pm 0.033$ \\
& STaylor  & $0.720 \pm 0.140$ & $0.934 \pm 0.011$ & $0.840 \pm 0.207$ & $0.160 \pm 0.207$ & $0.232 \pm 0.032$ & $0.841 \pm 0.031$ & $0.190 \pm 0.043$ \\
\midrule
\multirow{6}{*}{Bike}
& SHAP     & $1.000 \pm 0.000$ & $0.587 \pm 0.063$ & $0.720 \pm 0.193$ & $0.280 \pm 0.193$ & $0.267 \pm 0.155$ & $0.832 \pm 0.089$ & $0.216 \pm 0.145$ \\
& CondSHAP & $0.620 \pm 0.114$ & $0.589 \pm 0.067$ & $0.760 \pm 0.126$ & $0.240 \pm 0.126$ & $0.169 \pm 0.100$ & $0.895 \pm 0.055$ & $0.122 \pm 0.080$ \\
& SAGE     & $0.760 \pm 0.084$ & $0.615 \pm 0.068$ & $0.700 \pm 0.105$ & $0.300 \pm 0.105$ & $0.378 \pm 0.230$ & $0.782 \pm 0.102$ & $0.299 \pm 0.171$ \\
& ShapEff  & $0.680 \pm 0.140$ & $0.584 \pm 0.071$ & $0.820 \pm 0.114$ & $0.180 \pm 0.114$ & $0.196 \pm 0.116$ & $0.873 \pm 0.059$ & $0.151 \pm 0.085$ \\
& BShap    & $0.760 \pm 0.126$ & $0.593 \pm 0.063$ & $0.800 \pm 0.211$ & $0.200 \pm 0.211$ & $0.251 \pm 0.163$ & $0.849 \pm 0.089$ & $0.192 \pm 0.143$ \\
& STaylor  & $0.620 \pm 0.148$ & $0.528 \pm 0.109$ & $0.780 \pm 0.148$ & $0.220 \pm 0.148$ & $0.177 \pm 0.077$ & $0.870 \pm 0.072$ & $0.157 \pm 0.112$ \\
\midrule
\multirow{6}{*}{CAH}
& SHAP     & $1.000 \pm 0.000$ & $0.838 \pm 0.026$ & $0.940 \pm 0.097$ & $0.060 \pm 0.097$ & $0.170 \pm 0.033$ & $0.936 \pm 0.012$ & $0.068 \pm 0.013$ \\
& CondSHAP & $0.680 \pm 0.169$ & $0.792 \pm 0.044$ & $0.800 \pm 0.163$ & $0.200 \pm 0.163$ & $0.208 \pm 0.088$ & $0.919 \pm 0.020$ & $0.088 \pm 0.023$ \\
& SAGE     & $0.780 \pm 0.114$ & $0.800 \pm 0.042$ & $0.780 \pm 0.148$ & $0.220 \pm 0.148$ & $0.158 \pm 0.052$ & $0.925 \pm 0.029$ & $0.082 \pm 0.036$ \\
& ShapEff  & $0.740 \pm 0.097$ & $0.784 \pm 0.033$ & $0.800 \pm 0.133$ & $0.200 \pm 0.133$ & $0.158 \pm 0.069$ & $0.930 \pm 0.008$ & $0.075 \pm 0.009$ \\
& BShap    & $0.920 \pm 0.103$ & $0.844 \pm 0.016$ & $1.000 \pm 0.000$ & $0.000 \pm 0.000$ & $0.169 \pm 0.026$ & $0.938 \pm 0.015$ & $0.066 \pm 0.017$ \\
& STaylor  & $0.880 \pm 0.103$ & $0.837 \pm 0.014$ & $0.960 \pm 0.084$ & $0.040 \pm 0.084$ & $0.166 \pm 0.027$ & $0.936 \pm 0.010$ & $0.069 \pm 0.012$ \\
\midrule
\multirow{6}{*}{CCS}
& SHAP     & $1.000 \pm 0.000$ & $0.663 \pm 0.028$ & $1.000 \pm 0.000$ & $0.000 \pm 0.000$ & $0.149 \pm 0.008$ & $0.928 \pm 0.010$ & $0.077 \pm 0.012$ \\
& CondSHAP & $0.700 \pm 0.105$ & $0.546 \pm 0.092$ & $0.880 \pm 0.140$ & $0.120 \pm 0.140$ & $0.169 \pm 0.027$ & $0.899 \pm 0.021$ & $0.112 \pm 0.027$ \\
& SAGE     & $0.700 \pm 0.141$ & $0.635 \pm 0.054$ & $1.000 \pm 0.000$ & $0.000 \pm 0.000$ & $0.164 \pm 0.019$ & $0.907 \pm 0.017$ & $0.102 \pm 0.021$ \\
& ShapEff  & $0.740 \pm 0.135$ & $0.646 \pm 0.046$ & $0.980 \pm 0.063$ & $0.020 \pm 0.063$ & $0.160 \pm 0.013$ & $0.908 \pm 0.013$ & $0.101 \pm 0.016$ \\
& BShap    & $0.880 \pm 0.103$ & $0.676 \pm 0.045$ & $1.000 \pm 0.000$ & $0.000 \pm 0.000$ & $0.152 \pm 0.013$ & $0.931 \pm 0.015$ & $0.074 \pm 0.018$ \\
& STaylor  & $0.740 \pm 0.165$ & $0.639 \pm 0.055$ & $0.980 \pm 0.063$ & $0.020 \pm 0.063$ & $0.159 \pm 0.019$ & $0.917 \pm 0.022$ & $0.091 \pm 0.026$ \\
\midrule
\multirow{6}{*}{EE}
& SHAP     & $1.000 \pm 0.000$ & $0.948 \pm 0.012$ & $0.820 \pm 0.175$ & $0.180 \pm 0.175$ & $0.200 \pm 0.082$ & $0.911 \pm 0.024$ & $0.098 \pm 0.030$ \\
& CondSHAP & $0.700 \pm 0.105$ & $0.936 \pm 0.030$ & $0.800 \pm 0.163$ & $0.200 \pm 0.163$ & $0.174 \pm 0.077$ & $0.893 \pm 0.085$ & $0.133 \pm 0.141$ \\
& SAGE     & $0.880 \pm 0.103$ & $0.939 \pm 0.027$ & $0.940 \pm 0.097$ & $0.060 \pm 0.097$ & $0.276 \pm 0.120$ & $0.901 \pm 0.038$ & $0.111 \pm 0.049$ \\
& ShapEff  & $0.880 \pm 0.140$ & $0.935 \pm 0.037$ & $0.920 \pm 0.103$ & $0.080 \pm 0.103$ & $0.275 \pm 0.124$ & $0.899 \pm 0.037$ & $0.115 \pm 0.048$ \\
& BShap    & $0.840 \pm 0.084$ & $0.949 \pm 0.010$ & $0.900 \pm 0.141$ & $0.100 \pm 0.141$ & $0.212 \pm 0.089$ & $0.905 \pm 0.019$ & $0.105 \pm 0.024$ \\
& STaylor  & $0.780 \pm 0.148$ & $0.935 \pm 0.036$ & $0.920 \pm 0.103$ & $0.080 \pm 0.103$ & $0.258 \pm 0.129$ & $0.890 \pm 0.041$ & $0.126 \pm 0.053$ \\
\bottomrule
\end{tabular}
}
\end{table}

\subsection{Summary of Sensitivity Analyses}
\label{app:sensitivity_summary}

The sensitivity analyses support three conclusions about the proposed table. First, the main diagnostics are not driven by a single table size or explanation-sample choice. Increasing \(k\) often improves surrogate fidelity, but the gains do not automatically reduce ambiguity or improve coefficient stability. Larger explanation samples mainly sharpen coefficient estimates by narrowing intervals and reducing bootstrap dispersion, while the structural adequacy of the selected-feature surrogate remains dataset-dependent.

Second, the SHAP-ranking sample size and black-box backbone affect the upstream explanation problem in different ways. The SHAP top-\(k\) ranking is fairly stable across the sampled ranges, so downstream diagnostic changes are limited when the selected feature set changes little. By contrast, changing the black-box backbone changes the fitted response being explained, and therefore changes the resulting \(\phi\)-table diagnostics. This is expected: the table is designed to summarize \(f(X)\), so robustness means that the construction remains applicable across fitted models, not that different black boxes produce identical explanations.

Third, replacing the SHAP ranker with other Shapley-based global rankers shows that the \(\phi\)-table is not tied to a particular SHAP implementation. Different rankers can select different feature sets, but once a feature set is selected, the same statistical table provides a common diagnostic layer for direction, uncertainty, fidelity, and coefficient stability. Overall, the appendix results indicate that the proposed table is best viewed as a stable construction applied after feature selection, while its substantive interpretation remains conditional on the selected features, the explanation sample, and the fitted model response.

\end{document}

%% file: math_commands.tex
\usepackage{amsmath,amsfonts,bm}

\def\eqref#1{equation~\ref{#1}}

\def\1{\bm{1}}

\DeclareMathAlphabet{\mathsfit}{\encodingdefault}{\sfdefault}{m}{sl}
\SetMathAlphabet{\mathsfit}{bold}{\encodingdefault}{\sfdefault}{bx}{n}

